\pdfoutput=1
\documentclass[letterpaper, 10 pt, conference]{ieeeconf}
\IEEEoverridecommandlockouts

\usepackage[per-mode=symbol]{siunitx}
\usepackage[bottom]{footmisc}
\usepackage[font={small}, figurename={Fig.}, tablename={Tab.}]{caption}
\usepackage[normalem]{ulem}
\usepackage[scaled=.8]{beramono}
\usepackage[scaled=0.86]{helvet}
\usepackage[utf8]{inputenc}

\usepackage{amsmath,amssymb,amsfonts, mathtools, amsthm}
\usepackage{balance}
\usepackage{bm}
\usepackage{booktabs}
\usepackage{censor}
\usepackage{color,soul}
\usepackage{dblfloatfix}
\usepackage{etoolbox}
\usepackage{floatrow}
\usepackage{float}
\usepackage{graphicx}
\usepackage{ifthen}
\usepackage{import}
\usepackage{listings}
\usepackage{marvosym}
\usepackage{mathtools}
\usepackage{mdframed}
\usepackage{multirow}
\usepackage{nicefrac}
\usepackage{pict2e}
\usepackage{subcaption}
\usepackage{svg}
\usepackage{tabularx}
\usepackage{textcomp}
\usepackage{xcolor}
\usepackage{xspace}
\usepackage[percent]{overpic}
\usepackage[style=ieee, backend=biber, bibencoding=utf8, maxbibnames=10, mincitenames=1, maxcitenames=2, url=false, doi=false, isbn=false, citestyle=ieee, natbib=true, eprint=true, alldates=year]{biblatex}
\usepackage[acronym, nohypertypes={acronym}]{glossaries}

\floatstyle{plaintop}

\newcommand{\vect}[1]{\boldsymbol{\mathbf{#1}}}
\restylefloat{table}
\DeclareSourcemap{
  \maps[datatype=bibtex, overwrite]{
    \map{
        \step[fieldset=address, null]
        \step[fieldset=location, null]
        \step[fieldset=editor, null]
        \step[fieldset=series, null]
        \step[fieldsource=issue, match=\regexp{\A[0-9]+\Z}, final]
        \step[fieldset=number, origfieldval]
        \step[fieldset=issue, null]
    }
    \map{
        \pertype{inproceedings}
        \step[fieldset=publisher, null]
        \step[fieldset=volume, null]
    }
  }
}

\makeatletter
\newcommand{\labeltext}[2]{%
  \@bsphack
  \csname phantomsection\endcsname % in case hyperref is used
  \def\@currentlabel{#1}{\label{#2}}%
  \@esphack
}
\makeatother

\PassOptionsToPackage{hyphens}{url}
\usepackage[hidelinks]{hyperref} 
\usepackage[capitalise, nameinlink]{cleveref}
\crefformat{equation}{(#2#1#3)}

\Crefname{figure}{Fig.}{Figs.}

\definecolor{britishracinggreen}{rgb}{0.0, 0.5, 0.1}
\definecolor{darkkkkgreen}{rgb}{0.0, 0.3, 0.0}
\definecolor{darkkkgreen}{rgb}{0.0, 0.66, 0.0}
\definecolor{darkkgreen}{rgb}{0.0, 0.78, 0.0}
\definecolor{darkblue}{rgb}{0.0, 0.15, 0.5}
\definecolor{purple1}{rgb}{1.0, 0.0, 0.55}
\definecolor{purple2}{rgb}{0.45, 0.2, 1.0}

\DeclareMathOperator\erf{erf}
\usepackage{color}
\definecolor{tumblue}{rgb}{0, 0.4, 0.74}
\definecolor{darkgreen}{rgb}{0.0, 0.5, 0.0}
\definecolor{lightgreen}{rgb}{0.0, 0.84, 0.0}
\newcommand\textlcsc[1]{\textsc{\MakeLowercase{#1}}}

\newcolumntype{L}[1]{>{\raggedright\arraybackslash}p{#1}}

\newcolumntype{C}[1]{>{\centering\arraybackslash}p{#1}}

\newcolumntype{R}[1]{>{\raggedleft\arraybackslash}p{#1}}

\renewcommand{\d}[1]{\ensuremath{\operatorname{d}\!{#1}}}

\title{\LARGE \bf  Deep Occupancy-Predictive Representations for Autonomous Driving}

\author{Eivind Meyer, Lars Frederik Peiss, and Matthias Althoff
    \thanks{Department of Informatics, Technical University of Munich, Garching, Germany. \url{{eivind.meyer, lf.peiss, althoff}@tum.de}}%
}

\begin{document}
\maketitle
% remove the two linesbelow to disable page count
\thispagestyle{plain}
\pagestyle{plain}

\begin{abstract}
Manually specifying features that capture the diversity in traffic environments is impractical. Consequently, learning-based agents cannot realize their full potential as neural motion planners for autonomous vehicles. Instead, this work proposes to \textit{learn} which features are task-relevant. Given its immediate relevance to motion planning, our proposed architecture encodes the probabilistic occupancy map as a proxy for obtaining pre-trained state representations of the environment. By leveraging a map-aware traffic graph formulation, our agent-centric encoder generalizes to arbitrary road networks and traffic situations.
We show that our approach significantly improves the downstream performance of a reinforcement learning agent operating in urban traffic environments.
\end{abstract}

%  Columnwidth: \the\columnwidth
%  Textwdith: \the\textwidth

\section{Introduction}

Human drivers inherently possess an ability to react to new situations. This is in stark contrast to the narrow operational domains of current reinforcement learning (RL) approaches for self-driving, due to the prevalence of ad hoc feature vectors associated with poor generalization~\cite{kiran_survey, tampuuSurveyEnd2End}. In particular, two domain-specific characteristics of autonomous driving render the systematic design of relevant and comprehensive state representations difficult: First, the variable number and lack of a canonical ordering of other traffic participants is incompatible with fixed-sized feature vectors. Second, the diversity in road networks in terms of geospatial topology complicates specifying a universal map representation~\cite{JIANG2007647}.

By adopting graph neural networks (GNNs) as RL policies, recent works have outperformed traditional approaches relying on fixed-sized feature vectors. However, these were confined to homogeneous road network geometries such as highways~\cite{huegle_dynamic_2019, hart_graph_2020} or roundabouts~\cite{ha_road_neural_networks_2021}, simplifying the learning problem. Recently, GNN architectures that unify traffic and infrastructure have been proposed for the related task of vehicle trajectory prediction~\cite{liang_lane_graph_2020, zeng_lanercnn_2021, kim_lapred_2021, janjo2021starnet, gilles2021gohome, MoHXL22}.
However, directly adopting heterogeneous GNNs as policy networks is challenging, as current state-of-the-art RL algorithms cannot be reliably trained in complex environments
~\cite{henderson2017deep, busoniu_reinforcement_2018, dulac2021challenges}.

\begin{figure}
\centering
\vspace{0.3cm}\hspace*{0.8cm}\begin{subfigure}[!htb]{\linewidth}
  \centering   
  \begin{overpic}[width=1.0\linewidth, trim=0 200 0 300,clip]{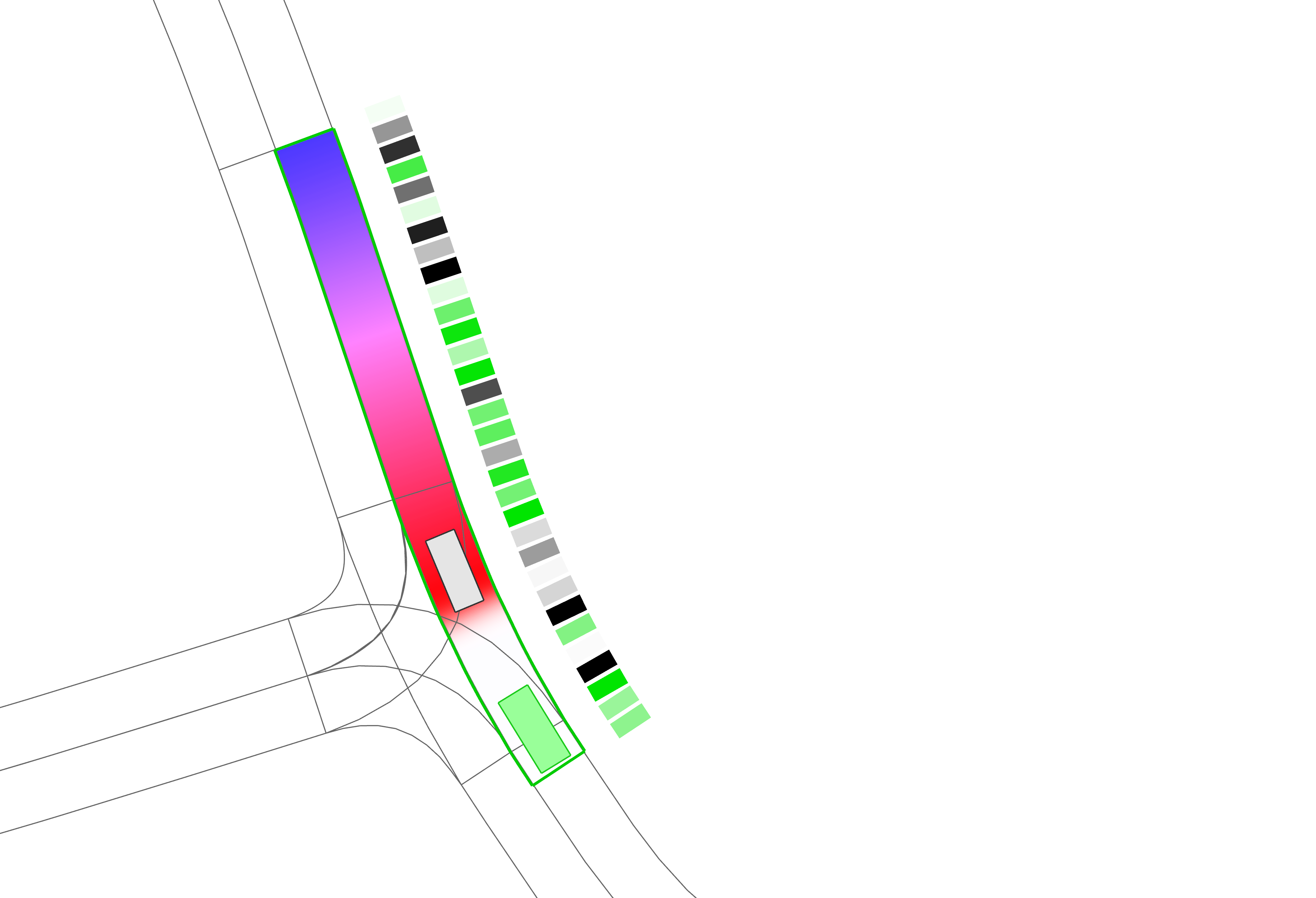}
         \put(29,48){
         \includegraphics[width=5.0cm]{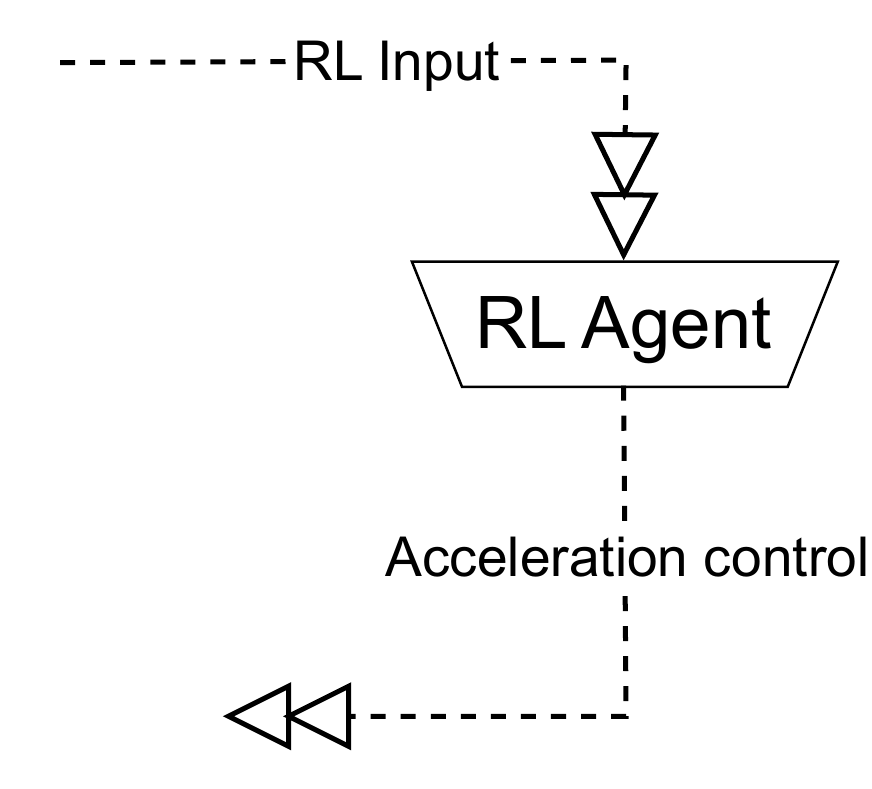}}
         \put (43,44) {\scriptsize{Pre-trained}}
         \put (43,40) {\scriptsize{latent states}}
         \put (50,38) {\color{black}\vector(-3, -1){10.7}}
         
         \put (-8.5,26) {\scriptsize{Spatio-temporal}}
         \put (-8.5,22) {\scriptsize{probabilistic occupancy map}}
         \put (-8.5,18) {\scriptsize{(human-interpretable)}}
         
         \put (9,35) { \small {$\hat{o}(s, t)$}}
         \put (41.5,32.5) { \small {$\vect{z}_{\mathrm{ego}}$}}
         \put (6,29.5) {\color{black}\vector(3, 1){19.2}}
         \put (18,0.8) {\scriptsize{Ego vehicle}}
         \put (25,4) {\color{black}\vector(4, 1){12.2}}
  \end{overpic}
\end{subfigure}
  \caption{
  Our spatio-temporal representation model learns a continuous parameterization of the probabilistic occupancy map $\hat{o}(s, t)$. The red and blue coloring scheme signifies high occupancy probability in the short and long-term future, respectively. The pre-trained intermediate states $\vect{z}_{\mathrm{ego}}$ are extracted as inputs for an RL agent controlling the longitudinal acceleration of the ego vehicle.
  \label{fig:occlandscapepage1}}
\end{figure}

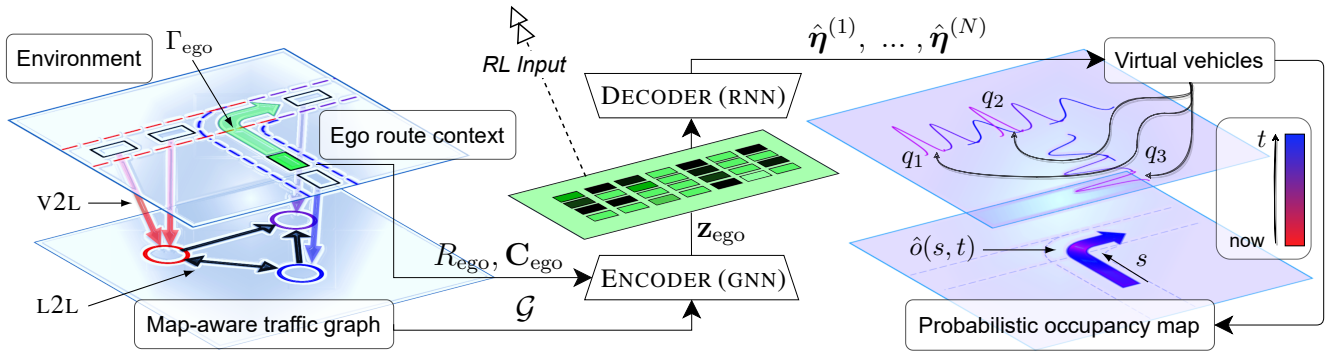
\begin{figure*}
   \vspace{0.1cm}\begin{overpic}[width=1.0\textwidth,tics=10]{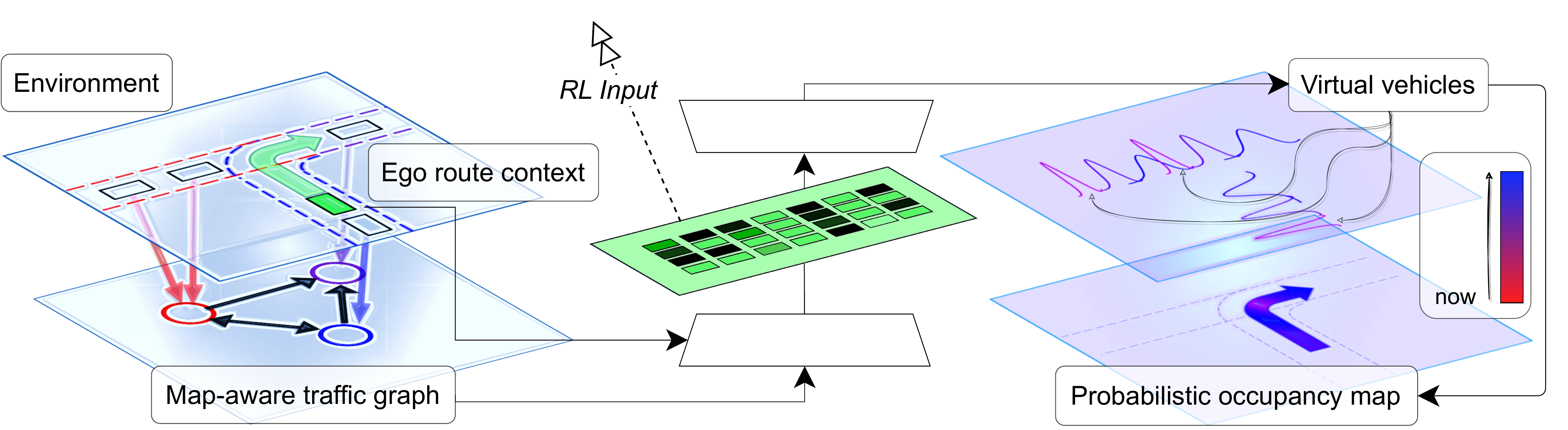}
 \put (38.15,2.67) {\large$\mathcal{G}$}
 \put (44.5, 5.0) {\normalsize$ \textsc{Encoder}\, \textlcsc{(gnn)}$}
 \put (44.5, 18.7) {\normalsize$ \textsc{Decoder}\, \textlcsc{(rnn)}$}
 \put (32,6.7) {\large$R_{\mathrm{ego}}, \vect{C}_{\mathrm{ego}}$}
 \put (51.7,8.8) {\large$\vect{z}_{\mathrm{ego}}$}
 \put (60,23.0) {\large$\hat{\vect{\eta}}^{(1)}, \; ... \;, \hat{\vect{\eta}}^{(N)}$}
 \put (67.6, 7.4) {\rotatebox{0} {\normalsize$\hat{o}(s, t)$}}
 \put(73.1, 7.8){\color{black}\vector(1.0,0){5.5}}
 \put (84.5, 6.6) {\rotatebox{0} {\normalsize$s$}}
 \put(85.45, 5.4){\color{black}\vector(-1.6,1.0){3.5}}
 
 \put (93.6,15.5) {\normalsize$t$}
 \put (67,14.3) {\normalsize$q_1$}
 \put (73,18.8) {\normalsize$q_2$}
 \put (85.2,14.7) {\normalsize$q_3$}
 \put (2.3, 10.8) {\rotatebox{0.0} {\normalsize$\textsc{v2l}$}}
 \put(6.2, 11.3){\color{black}\vector(1.0,0){3.3}}
  \put (2.3,3.05) {\rotatebox{0.0} {\normalsize$\textsc{l2l}$}}
 \put(6.2,3.6){\color{black}\vector(9.0,3){8}}
 \put (12.0,22.8) {\color{black}\normalsize$\Gamma_{\mathrm{ego}}$}
 \put(14,22){\color{black}\vector(1.2,-2){3.05}}
\end{overpic}
 \caption{Overview of our proposed architecture: The fixed-sized latent states $\vect{z}_{\mathrm{ego}}$, which are encoded by an ego-conditioned, heterogeneous graph neural network, are extracted as inputs for an RL-based motion planner. Our novel decoder architecture, which infers the probabilistic occupancy map from reconstructed virtual vehicles, is used for pre-training the encoder and can be disabled during inference.}
 \label{fig:overview}
\end{figure*}

To mitigate the challenging nature of the learning task, we instead formulate a representation objective and design a state representation model detached from the RL training loop. As opposed to letting the agent directly infer control signals from a multi-modal graph representation, we use spatio-temporal occupancy map prediction, as depicted in~\cref{fig:occlandscapepage1}, as a learning proxy for environment \textit{understanding}. As illustrated in~\cref{fig:overview}, we specifically develop a GNN-based encoder-decoder model whose intermediate latent states serve as low-dimensional, pre-trained state representations. Notably, the flexible nature of our encoder allows arbitrary road network topologies and traffic environments to be captured by the learned representations. To alleviate the lossy nature of compressive graph encoding, we propose a novel occupancy prediction framework that constrains the decoding space in accordance with a priori known physical priors for vehicle motion. We implement our approach using CommonRoad-Geometric (\textit{crgeo})~\cite{meyer2023geometric}, a PyTorch-based framework offering a standardized graph extraction pipeline for traffic scenarios. Our source code is available at {\small\url{https://github.com/CommonRoad/crgeo-learning}}.
\section{Related work}\label{sec:background}

We first introduce the learning frameworks used by our approach alongside related applications to motion planning.

\subsection{State representation learning (SRL)}

By learning encoded representations of the surroundings, SRL methods enhance the performance of RL agents operating in high-dimensional, complex environments~\cite{munk_2016, lesort_state_2018, schwarzer_pretraining_2021}.
An \textit{agent-centric} approach is generally preferred, so that the learned representations are aligned with the planning context~\cite{parisi_2017}. Further desired requirements for representations are they enable predicting the future world state~\cite{boots_2011_predictive, guo2020bootstrap, lee2020predictive, recanatesi_2021,  ha_recurrent_2018} (as opposed to merely reconstructing the present) and that they are \textit{low-dimensional}~\cite{nouri_dimension_2010, pmlr-v83-bassily18a}. To mitigate the trade-off between dimensionality reduction and expressiveness, SRL can be supported by incorporating knowledge about the world as \textit{representation priors}~\cite{lesort_state_2018, jonschkowski_learning_2015, bengio_representation_2014}. Imposing structural constraints on the representations, e.g. by enforcing correspondence to physically plausible world states, improves their generalization and downstream effectiveness~\cite{Scholz2014APM, stewart2016labelfree}.

\subsection{Graph neural networks (GNNs)}
As the graph-compatible counterpart of traditional encoder architectures, GNNs present a framework for applying SRL to traffic environments. 
Within the context of the widely adopted \textit{message passing} paradigm~\cite{gilmer2017neural}, GNNs compute neighborhood-aware hidden representations via the permutation-invariant aggregation of \textit{edge messages}, i.e., neural encodings transmitted from a node to its (outgoing) neighbors. This facilitates the propagation of task-relevant information flow across the graph, which, depending on the learning problem, can be summarized on the graph level via readout operations~\cite{zhou2018graph}. As necessitated by the multi-modal traffic graph formulation assumed in this work, GNNs are also extendable to heterogeneous graph inputs~\cite{schlichtkrull_2018, hetero_gnn_wang_2019}.

\subsection{Applications to motion planning}
Autoencoder-based representation models~\cite{tschannen2018recent} have been used in a multitude of existing works for learning latent states based on, e.g., rasterized bird's eye view images~\cite{toghi2021learning, sama_2018, zhao_scenario_rep_2021, kendall_2019, dignet_2020_cai} or on-board sensor data~\cite{dong_spatio-weighted_2020}. However, they do not leverage the structural biases~\cite{battaglia_2018} induced by the road network topology. In line with our approach,~\cite{sanchez2018graph} uses a GNN-based encoder to learn structurally-aware state representations, but in the context of RL-based robotic manipulators. 

Road occupancy as a standalone prediction task is widely covered in existing works~\cite{hoermann_occpred_2018, mohajerin_occpred_2019, schreiber_occpred_2019, gilles_occpred_2021, kaniarasu_2021_goaldir_occ_pred}. With the objective of encoding traffic scenes similar to ours (albeit not in the context of motion planning), encoder architectures for learning representations of occupancy maps have been proposed~\cite{rakos_2021, marina_deep_2019, itkina_2019}. Using graphical or otherwise spatially-aware encoders similar to ours, recent works such as~\cite{amirloo_2021, hu_freespace_2021, casas_2021, sadat_2020, mahjourian_occupancy_flow_2022} predict occupancy grids~\cite{elfes_1989} as an intermediate learning target for guiding the training of neural motion planners. However, these approaches do not provide global,
low-dimensional representations appropriate for decoupled
RL agents. In contrast to our work, they also suffer from the lossy nature of grid-wise occupancy
discretization \cite{ilievski_design_space}.
\section{Methodology}\label{sec:representation}

Next, we outline the details of our approach.

\subsection{Definitions}

\subsubsection{Heterogeneous traffic graph}
As originally proposed in~\cite{bender_lanelets}, we model road networks as atomic, interconnected road segments (i.e., \textit{lanelets}). We formalize the dynamic traffic environment at the current time step by the heterogeneous graph tuple ${\mathcal{G} = (\mathcal{V}, \mathcal{E}, X_\mathcal{V}, X_\mathcal{E})}$, where ${\mathcal{V} = (\mathcal{V}_\textsc{v}, \mathcal{V}_{\textsc{l}})}$ indexes the vehicle ($\textsc{v}$) and lanelet ($\textsc{l}$) nodes, ${\mathcal{E} = (\mathcal{E}_{\textsc{v2l}}, \mathcal{E}_{\textsc{l2l}})}$ defines the corresponding vehicle-to-lanelet ($\textsc{v2l}$) and lanelet-to-lanelet ($\textsc{l2l}$) edges, and ${X_\mathcal{V} = (\vect{X}_{\textsc{v}}, \vect{X}_{\textsc{l}})}$ and ${X_\mathcal{E} = (\vect{X}_{\textsc{v2l}}, \vect{X}_{\textsc{l2l}})}$ contain node and edge-level graph features, respectively. Here, the time-dependent $\textsc{v2l}$ edges relate to the physical presence of a vehicle on a given lanelet, whereas the static $\textsc{l2l}$ edges are implied by the road network topology.
Our approach incorporates the default graph features provided by \textit{crgeo}~\cite{meyer2023geometric}, e.g. velocity ($\textsc{v}$) and vehicle-lanelet heading difference ($\textsc{v2l}$). 

\subsubsection{Planning context}
\begin{figure}[t!]
{\caption{Ego reference path $\Gamma_{\mathrm{ego}}$ composed of two successive lanelets given by $R_{\mathrm{ego}} = [1, 2]$.
}
    \label{fig:ego_ref_path}}
\vspace{0.05cm}{\begin{overpic}[width=2.8cm,angle=-90,tics=10, trim=59 0 70 0,clip]{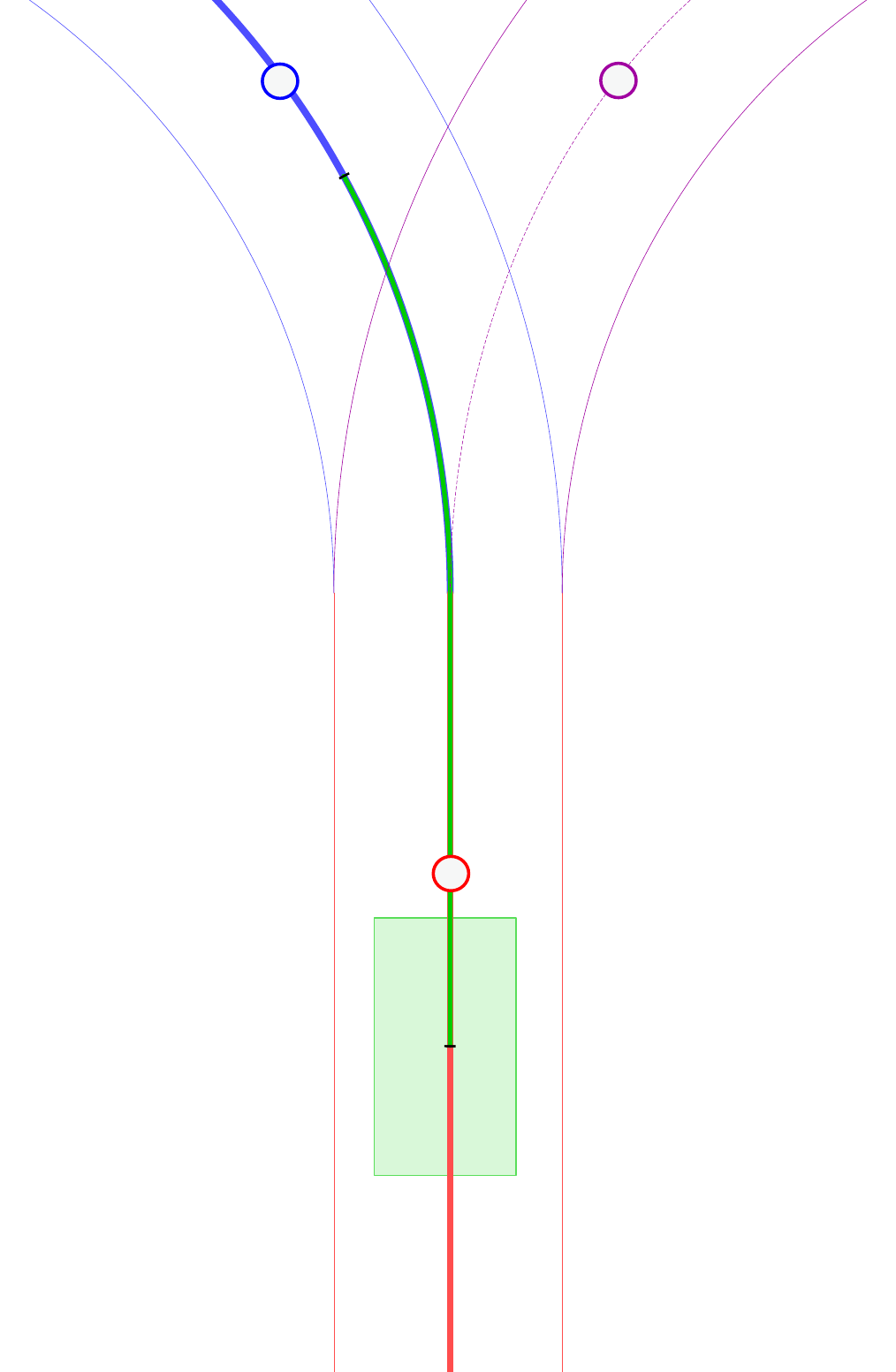}
 \put (19.78, 4.2) {\scriptsize$\displaystyle\color{red}{s^{\mathrm{start}}_1}$} 
 \put(23.6,  9.5){\color{red}\vector(0, 1){6.5}}
 \put (43.4,27) {\footnotesize$\displaystyle\color{darkkkgreen}\Gamma_{\mathrm{ego}}$} 
\put(46.4,  24.5){\color{darkkgreen}\vector(0, -1){7.5}}
 \linethickness{0.5pt}
 \put (35.5, 12.2) {\scriptsize$\displaystyle\color{black}1$}
 \put (96.2, 26.8) {\scriptsize$\displaystyle\color{black}2$}
 \put (96.1, 5.7) {\scriptsize$\displaystyle\color{black}3$}
 \put (84.3, 11.2) {\scriptsize$\displaystyle\color{blue}s^{\mathrm{end}}_2$}
 \put(87.3,  16.3){\color{blue}\vector(0, 1){7.0}}
\end{overpic} }
\end{figure}
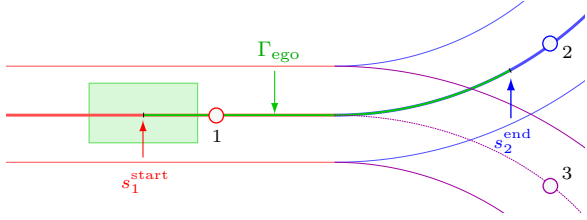

The state of the ego vehicle is chosen as the tuple $(\vect{p}_{\mathrm{ego}}, v_{\mathrm{ego}})$, consisting of its x-y center position and longitudinal speed. We further denote its length as $\lambda_{\mathrm{ego}}$.
Next, we let the reference path ${\Gamma_{\mathrm{ego}} : [0, \zeta_{\mathrm{ego}}] \rightarrow \mathbb{R}^2}$ of length $\zeta_{\mathrm{ego}}$ be parameterized by arclength $s$, and impose the natural constraint that ${\Gamma_{\mathrm{ego}}(0) = \vect{p}_{\mathrm{ego}}}$.
As illustrated in~\cref{fig:ego_ref_path}, we assume that ${\Gamma_{\mathrm{ego}}}$ follows the centerline of a connected, traffic-compliant sequence of lanelets. The corresponding sequence of lanelet node indices in ${\mathcal{V}_{\textsc{l}}}$ is denoted by $R_{\mathrm{ego}}$. Further, we let $s^{\mathrm{start}}_{j}$ and $s^{\mathrm{end}}_{j}$ denote the start and endpoint coordinates of the $j\textsuperscript{th}$ element in $R_{\mathrm{ego}}$, as defined within the arclength-parameterized coordinate frame of the centerlines. Also, we let $d_{j}$ denote lanelet length, and let $d_{j}^{\mathrm{prior}}$ be the aggregated length of the path segments preceding $j$. Finally, we let the spatial context matrix $\vect{C}_{\mathrm{ego}}$ contain the row vectors ${\vect{c}_{j} = [s^{\mathrm{start}}_{j},\, s^{\mathrm{end}}_{j},\,  d_{j},\,  d_{j}^{\mathrm{prior}}]}$. 

\subsection{Occupancy as representation objective}
As occupancy explicitly expresses drivable and non-drivable space, it can be considered as \textit{the} foundational environment characteristic in a motion planning context \cite{comp_sontges}. We consider occupancy solely in the longitudinal direction, as this is more relevant than the lateral direction and simplifies our modelling assumptions. For a given spatio-temporal coordinate vector $[s, t] \in \mathbb{R}^2$, future path occupancy on $\Gamma_{\mathrm{ego}}$ can be formalized as ${o : [0, \zeta_{\mathrm{ego}}] \times \mathbb{R} \rightarrow \{0, 1\}}$.
As shown in~\cref{fig:path_projection}, $o(s, t)$ is derived from a path projection of the vehicles that overlap with the road surface at time $t$.

\subsection{Encoder architecture}\label{sec:encoding}

Our encoding pipeline is formalized as
\begin{equation}
	\begin{aligned}
		\vect{z}_{\mathrm{ego}} &= \textsc{Encoder}(\mathcal{G}, R_{\mathrm{ego}}, \vect{C}_{\mathrm{ego}}),
	\end{aligned}
\label{eq:highlevelenc}
\end{equation}
with the low-dimensional representations $\vect{z}_{\mathrm{ego}} \in \mathbb{R}^{Z}$ being the final output of the encoder.  
The GNN-based encoder is designed to facilitate the probabilistic propagation of traffic participants across the given lanelet network, which is used as a neural infrastructure for social message passing. Next, we outline the encoding steps. In general, we use $\Theta_\square$ to denote trainable, nonlinear functions, $\Sigma$ to denote a permutation-invariant aggregation operation, and assume the activation function $\rho$ to be applied after each step.

\subsubsection{Vehicle-to-lanelet}
Unlike the vehicle-to-vehicle paradigm proposed in e.g.~\cite{hart_graph_2020}, we capture $\textsc{v2v}$ interaction effects in a map-agnostic fashion by first embedding the vehicles onto the lanelet graph. Letting vehicle and lanelet nodes be indexed by $i$ and $j$, respectively, we compute the initial hidden lanelet states $\vect{h}_{j}^{(0)} \in \mathbb{R}^{H}$ as
\begin{equation*}
	\begin{aligned}
		\vect{h}_{j}^{(0)}= \Theta_{\textsc{l}}(\vect{x}_j) 
  \smashoperator{ \sum_{(i, j)\in \mathcal{E}_{\textsc{v2l}}} } \Theta_{\textsc{v2l}}([\vect{x}_i, \vect{x}_j, \vect{x}_{i \rightarrow j}]).
	\end{aligned}
\label{eq:v2l}
\end{equation*}
where the node and edge features $(\vect{x}_i$, $\vect{x}_j)$ and $\vect{x}_{i \rightarrow j}$ denote the corresponding row vectors in $X_\mathcal{V}$ and $X_\mathcal{E}$, respectively.
\subsubsection{Lanelet-to-lanelet}
Next, a total of $L$ successive $\textsc{l2l}$ message passing layers are used to facilitate the propagation of information flow across the lanelet network.
With the superscript $l$ denoting the layer index, we recursively update the hidden lanelet states according to the update equation
\begin{equation*}
	\begin{aligned}
		\vect{h}_{j}^{(l+1)} = \vect{h}_{j}^{(l)} + 
  \smashoperator{ \sum_{(j^{'}\!\!, j)\in \mathcal{E}_{\textsc{l2l}}} } \Theta_{\textsc{l2l}}([\vect{h}_{j^{'}}^{(l)}, \vect{h}_{j}^{(l)}, \vect{x}_{j^{'} \rightarrow j}]),
	\end{aligned}
\label{eq:l2l}
\end{equation*}

\begin{figure}
\vspace*{0.43cm}\begin{overpic}[width=2.4cm,angle=-90,tics=10, trim=0 0 150 0,clip]{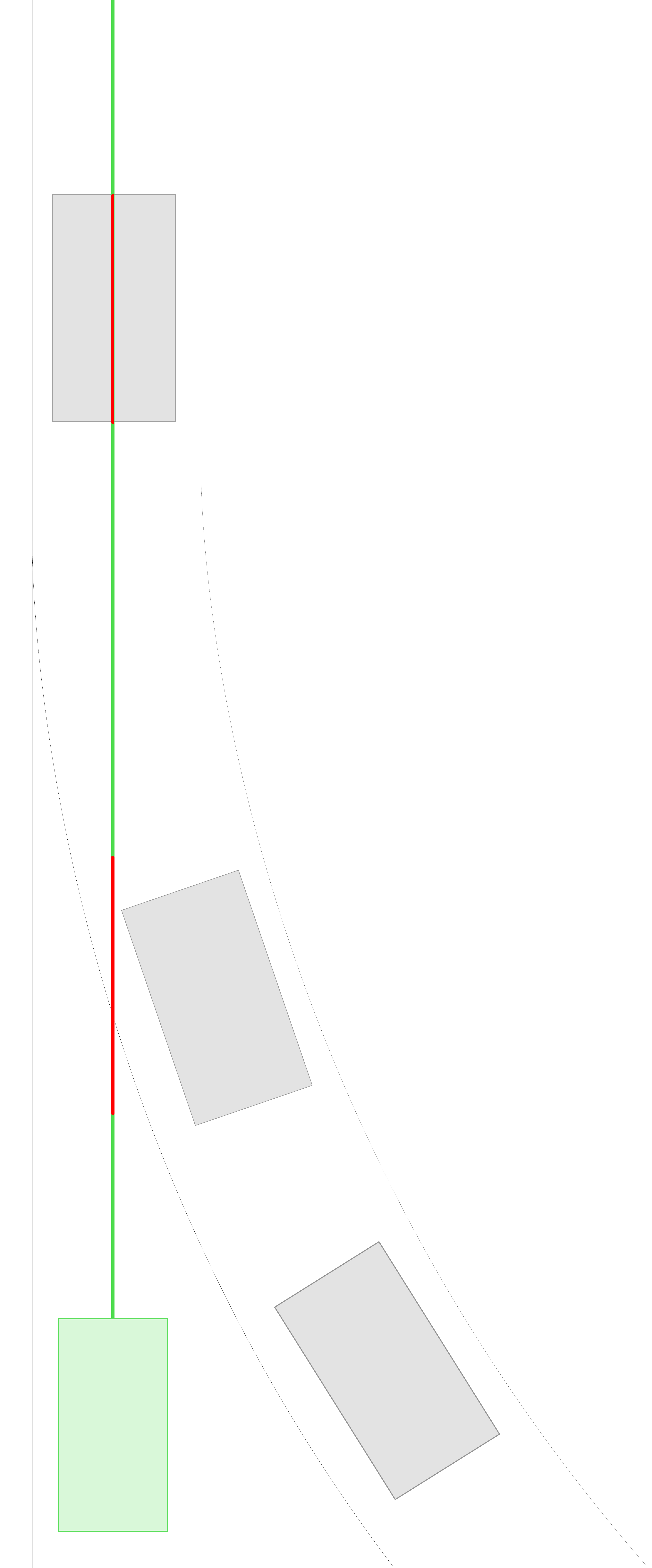}
{\caption{The ground-truth path occupancy $o$ is derived from projecting the vehicles in $\mathcal{V}_\textsc{v}$ onto $\Gamma_{\mathrm{ego}}$. The occupied and non-occupied path segments are colored red and green, respectively.}\label{fig:path_projection}}
 \put (0.1, 39) {\color{black}\scriptsize{Ego vehicle}}
 \put(9, 38){\color{black}\vector(0, -5){10}}
 \put (57, 39) {\color{black}\scriptsize{$\Gamma_{\mathrm{ego}}$}}
 \put(58.6, 38){\color{black}\vector(0, -5){10}}
 \put (45, 3) {\color{black}\scriptsize{Other traffic participants}}
 \put(57, 7){\color{black}\vector(4, 3){18}}
 \put(48, 7){\color{black}\vector(-3, 3){8}}
 \put(43.5, 4){\color{black}\vector(-8, 1){20}}
\end{overpic}
\end{figure}
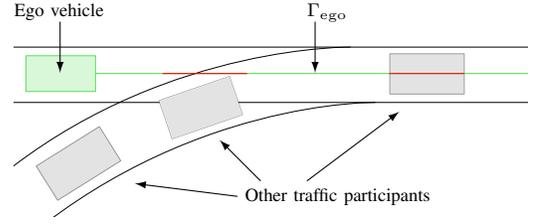

\subsubsection {Ego-attentional readout operation}
Then, an attentional readout layer~\cite{velickovic2018graph} is used to compute graph-level hidden states ${\vect{h}_{\mathrm{ego}} \in \mathbb{R}^{H}}$. To obtain ego-centered representations, the aggregation of $\vect{h}_{j}^{(L)}$ is weighted by attention scores $\vect{\alpha}_{\mathrm{ego}}$ based on the ego vehicle's spatial context $\vect{C}_{\mathrm{ego}}$, i.e., 
\begin{equation*}
\begin{aligned}
 \vect{\alpha}_{\mathrm{ego}} &= \mathrm{softmax}\left(\Theta_{C}(\vect{C}_{\mathrm{ego}})\right), \\
\vect{h}_{\mathrm{ego}} &= \sum_{j \in R_{\mathrm{ego}}}{\alpha_{\mathrm{ego}, j} \vect{h}_{j}^{(L)}}.
\end{aligned}
\end{equation*}
\subsubsection {Downscaling layer}
Finally, a downscaling MLP layer is applied to obtain the final latent states $\vect{z}_{\mathrm{ego}} = \Theta_z(\vect{h}_{\mathrm{ego}})$, intended to be used as state observations for the RL agent.
\subsection{Decoder architecture}\label{sec:decoding}
Our proposed decoder maps ${\vect{z}_{\mathrm{ego}}}$ to a continuous parameterization of the probabilistic occupancy map according to
\begin{equation}
\begin{aligned}
    \hat{o}(s, t) &= \textsc{Decoder}(\vect{z}_{\mathrm{ego}}, s, t), \\
    o(s, t) &\sim \mathrm{Bernoulli}(\hat{o}(s, t)).
\end{aligned}
\end{equation}\label{eq:highleveldecode}
However, as an abstraction of something tangible (i.e., the presence of vehicles), predicting $o(s, t)$ in an unconstrained fashion using, e.g., a regular MLP network, might lead to overparameterized predictions that are inconsistent with the data~\cite{thrun_learning_nodate}. The large hypothesis space is especially problematic given the low-dimensional nature of $\vect{z}_{\mathrm{ego}}$ and the resulting information bottleneck. Instead, our novel decoding architecture prevents nonsensical predictions of the occupancy map by inferring it from a decoded set of time-evolving probability distributions referred to as \textit{virtual vehicles}. This enforces temporal and spatial consistency on the output space and streamlines the learning task. By exploiting the physical priors of our application domain, our method further guarantees that the decoded occupancy maps conform to plausible limits for e.g. vehicle length and velocity. 
\subsubsection{Virtual vehicles}
We parameterize the probabilistic occupancy map $\hat{o}(s, t)$ via recurrently decoded virtual vehicles.
As outlined in~\cref{sec:encoding}, our fixed-sized intermediate representations $\vect{z}_{\mathrm{ego}}$ are computed by aggregating the elements in $\mathcal{G}$. Due to the resulting node-level data association loss, it is not viable to reconstruct vehicle instances in a way comparable to related trajectory prediction works~\cite{liang_lane_graph_2020, zeng_lanercnn_2021, kim_lapred_2021, janjo2021starnet, gilles2021gohome, MoHXL22}.  
Without assuming an association between decoded and actual vehicles, our virtual vehicle formulation instead enables a differentiable and permutation-invariant parameterization of the joint probabilistic occupancy map $\hat{o}(s, t)$. This translates the learning problem to the graph-level (i.e., global) domain, yielding a feasible training target. 

\subsubsection{Formal definition}
Formally, we let the state of a decoded virtual vehicle $q$ of length $\lambda_q \in \mathbb{R}$ be defined by the tuple ${(\mathcal{I}_q, p_q)}$, where ${\mathcal{I}_q : \mathbb{R} \rightarrow \{0, 1\}}$ is an existence indicator at time ${t}$, and ${p_q : \mathbb{R} \rightarrow \mathbb{R}}$ models its time-dependent longitudinal center position on $\Gamma_{\mathrm{ego}}$. Further, we let ${o_q : [0, \zeta_{\mathrm{ego}}] \times \mathbb{R} \rightarrow \{0, 1\}}$ return its path occupancy
\begin{equation}
\begin{aligned}
    o_q(s, t)=\begin{cases}
               1 & \text{if} \quad \mathcal{I}_q(t) = 1 \wedge |s - p_q(t)| < \frac{\lambda_q}{2} , \\
               0 & \text{otherwise}.
            \end{cases}
\end{aligned}\label{eq:vv_occ_def} 
\end{equation}
This expression differs from the vehicle occupancy $o$ considered in~\cref{fig:path_projection}, as virtual vehicles serve as non-deterministic, atomic proxies for modelling future occupancy flow.

\subsubsection{Stochastic formulation}

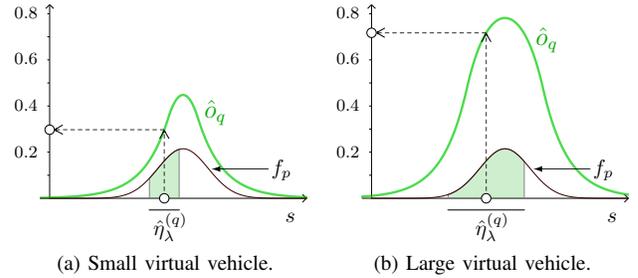
\begin{figure}[h!]%
    \centering
    \subfloat[Small virtual vehicle.
    \label{fig:narrowb}]{
    \resizebox{1.0\totalheight}{!}{
    \begin{overpic}[width=4.6cm,tics=10]{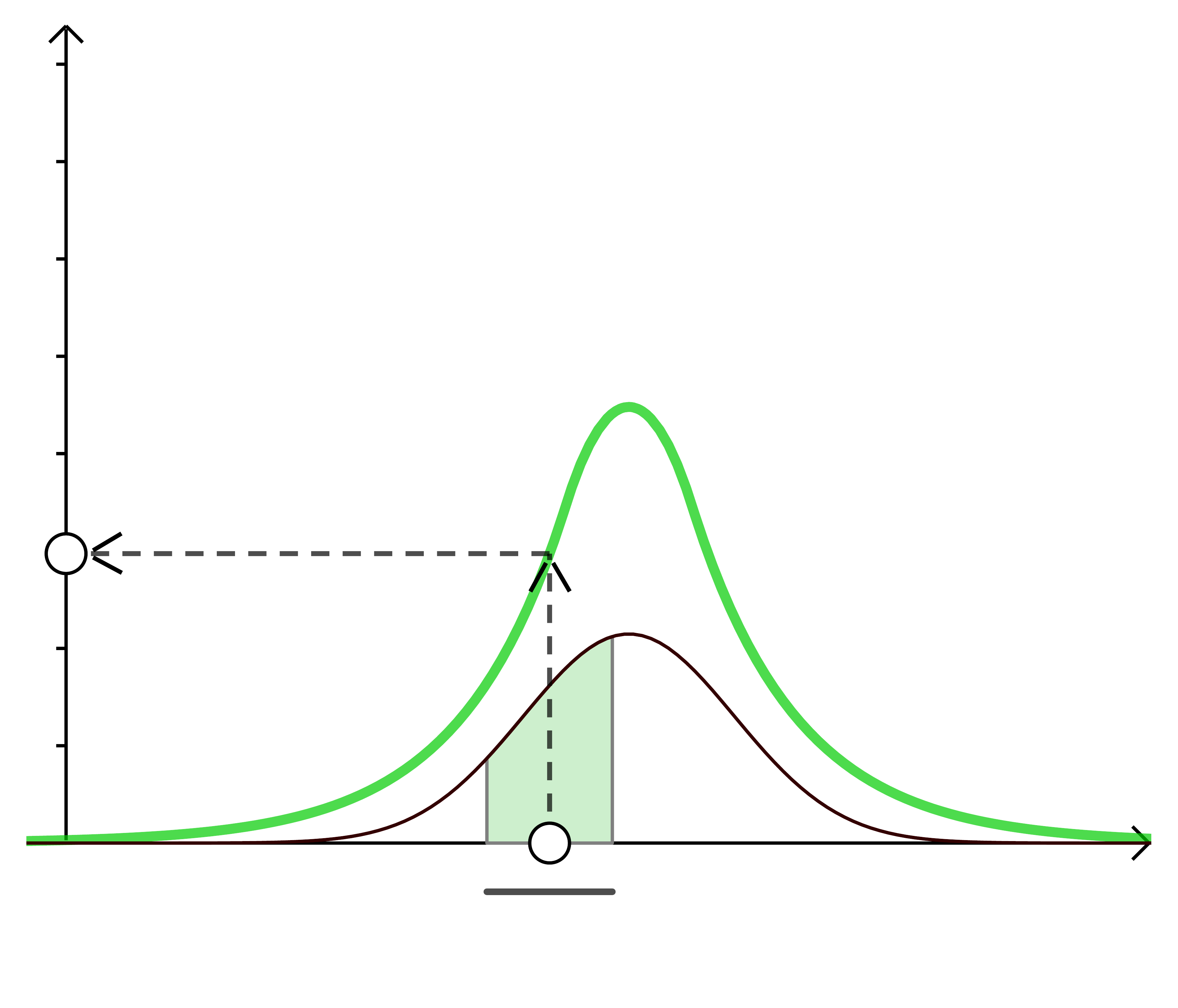}
 \put (42.8, 0.9) {\normalsize$\hat{\eta}_\lambda^{(q)}$}
 \put (90,5.5) {\normalsize$s$}
 \put (-7,28.5) {\footnotesize $0.2$}
 \put (-7,45.25) {\footnotesize $0.4$}
 \put (-7, 61.95) {\footnotesize $0.6$}
 \put (-7, 78.65) {\footnotesize $0.8$}
 \put (85,22.5) {\normalsize$f_p$}
 \put (61,43) {\color{darkkkgreen} \large$\hat{o}_q$}
 \put(84,24.5){\color{black}\vector(-1, 0){20.5}}
\end{overpic} 
}
    }%
    \subfloat[Large virtual vehicle.
    \label{fig:wideb}]{
    {
    \resizebox{1.0\totalheight}{!}{    
    \begin{overpic}[width=4.6cm,tics=10]{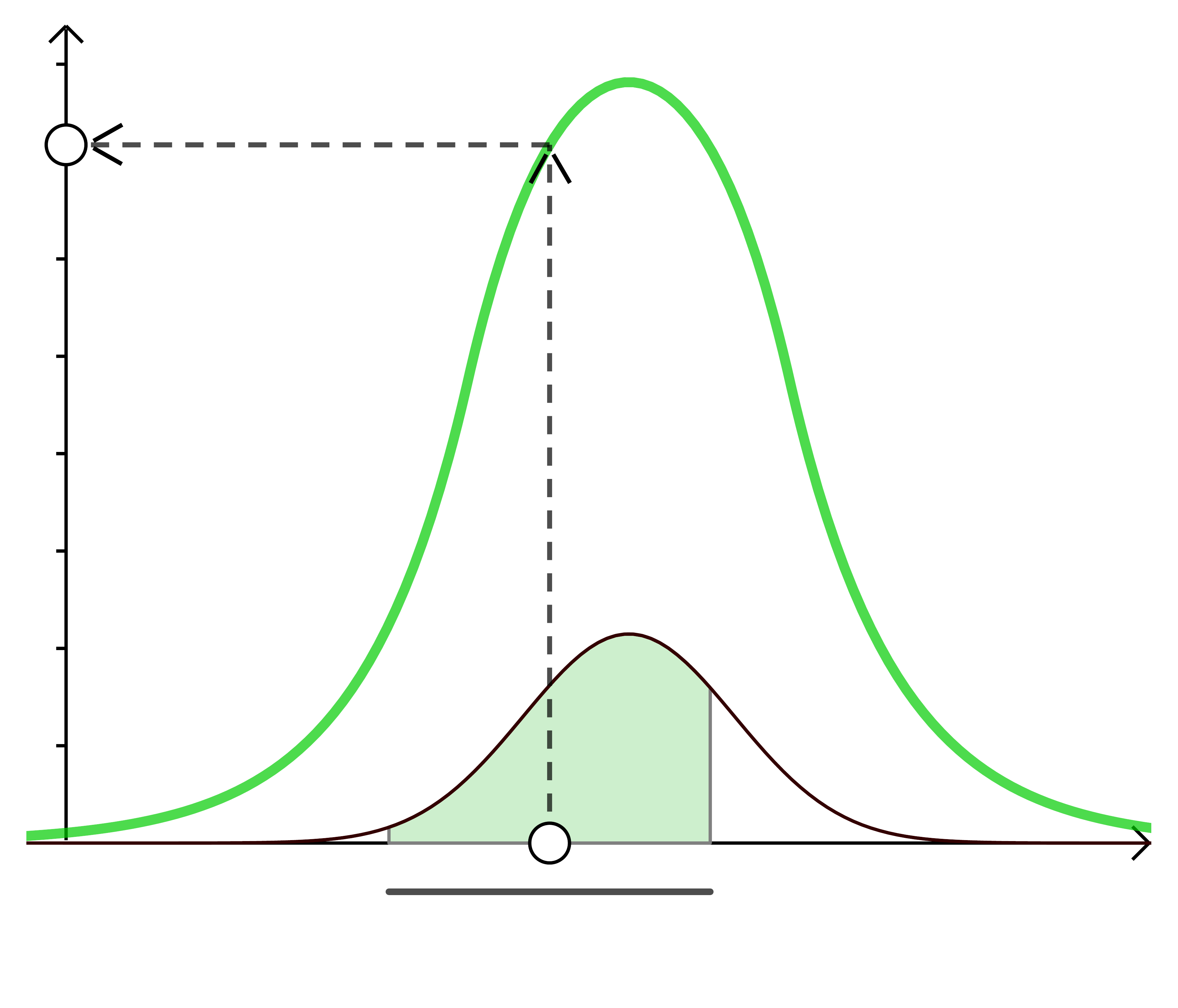}
 \put (42.8,0.9) {\normalsize$\hat{\eta}_\lambda^{(q)}$}
 \put (90,5.5) {\normalsize$s$}
 \put (-7,28.5) {\footnotesize $0.2$}
 \put (-7,45.25) {\footnotesize $0.4$}
 \put (-7, 61.95) {\footnotesize $0.6$}
 \put (-7, 78.65) {\footnotesize $0.8$}
 \put (85,22.5) {\normalsize$f_p$}
 \put (64.5,69) {\color{darkkkgreen} \large$\hat{o}_q$}
 \put(84,24.5){\color{black}\vector(-1, 0){20.5}}
\end{overpic}  
}
    }%
    }
 \caption{The predicted occupancy footprint $\hat{o}_q(s, t)$ of a virtual vehicle $q$ at a future time $t$ is derived via a symmetrically bounded integral of length $\hat{\eta}_\lambda^{(q)}$ centered at $s$. Here, the two vehicles have identical positional PDFs $f_p$, but different lengths $\hat{\eta}_\lambda^{(q)}$. As expected, a larger $\hat{\eta}_\lambda^{(q)}$ induces a larger predicted occupancy footprint.} 
    \label{fig:occpred}%
\end{figure}

We let $f_p : \mathbb{R} \times \mathbb{R} \rightarrow [0, 1]$ denote the time-varying probability density function (PDF) for $p_q$. As motivated in~\cite{hinkel2007applications}, the \textit{Fokker-Planck}~\cite{risken_fokkerplanck_1984} equation can be used for modelling microscopic traffic flow as a stochastic process. In this framework, $p_q$ is described by a partial differential equation influenced by stochastic forces, addressing the accumulation of uncertainty over time with regards to the vehicle position. As the model training requires a differentiable inference procedure, we use a simplified behavior model with a tractable solution. Specifically, we assume linear drift and diffusion terms given by $\hat{\eta}_{p, v}$ and $\hat{\eta}_{p, d}$, as well as a positional offset $\hat{\eta}_{p, \scriptscriptstyle{0}}$ corresponding to the initial vehicle position. Using ";" to separate the distributions' input space from their given parameterizations, this results in the time-evolving Gaussian solution~\cite{risken_fokkerplanck_1984}
\begin{equation}
\begin{aligned}
    f_p(s; t, \hat{\eta}_{p}) \!&=\! \frac{1}{\sqrt{4 \pi \hat{\eta}_{p, d} t}} \exp\!{\left(\!\!{-}\frac{(s - \hat{\eta}_{p, \scriptscriptstyle{0}} - \hat{\eta}_{p, v} t)^2}{4 \hat{\eta}_{p, d} t}\!\right)}, \\
    F_p(s; t, \hat{\eta}_{p}) \!&=\! - \frac{1}{2}
    \erf\! { \small{ \left(\! \dfrac{\hat{\eta}_{p, v} t + \hat{\eta}_{p, \scriptscriptstyle{0}} - s}{2 \sqrt{\hat{\eta}_{p, d}t}} \!\right)}},
\end{aligned}\label{eq:fokker_sol_pdf}
\end{equation}
with $F_p(s; t, \hat{\eta}_{p}) = \int_{-\infty}^{s} f_p(s'; t, \hat{\eta}_{p}^{(q)}) \d s'$ denoting the cumulative distribution function. Further, we let $f_{\mathcal{I}}: \mathbb{R} \rightarrow [0, 1]$ denote the predicted existence probability for $q$ at a given $t$. Delayed appearances or disappearances of vehicles on $\Gamma_{\mathrm{ego}}$ due to, e.g., lane changes cannot be captured by a constant existence probability $f_{\mathcal{I}}(t) = \hat{\eta}_{\mathcal{I}, \scriptscriptstyle{0}}$. By also introducing the temporal offset parameter $\hat{\eta}_{\mathcal{I}, \tau}$, we instead use the time-dependent parameterization given by
\begin{equation}
\begin{aligned}
    m_{l}(t, \hat{\eta}_{\mathcal{I}, \tau}) &= \sigma(\tau_R (t - \hat{\eta}_{\mathcal{I}, \tau} (1 \! + \! \tau_C) + \tau_C)), \\
    m_{r}(t, \hat{\eta}_{\mathcal{I}, \tau}) &= \sigma(\tau_R (1 - t + \hat{\eta}_{\mathcal{I}, \tau} (1 \! + \! \tau_C) + \tau_C)), \\
    m(t, \hat{\eta}_{\mathcal{I}, \tau}) &= m_{l}(t, \hat{\eta}_{\mathcal{I}, \tau})  \cdot m_{r}(t, \hat{\eta}_{\mathcal{I}, \tau}), \\
    f_{\mathcal{I}}(t; \hat{\eta}_{\mathcal{I}, \scriptscriptstyle{0}}, \hat{\eta}_{\mathcal{I}, \tau}) &= \hat{\eta}_{\mathcal{I}, \scriptscriptstyle{0}} \cdot m(t, \hat{\eta}_{\mathcal{I}, \tau}),
\end{aligned}\label{eq:mask}
\end{equation}
with $\sigma$ denoting the sigmoid function, and $\tau_C$ and $\tau_R$ being fixed model hyperparameters. As visualized in~\cref{fig:temporal_shifting}, the temporal masking effect produced by the left and right shifting mechanism $m$ provides the modelling flexibility for $f_{\mathcal{I}}$ to predict vehicles leaving or entering $\Gamma_{\mathrm{ego}}$ at future time instances.
Given the vector parameterization  ${\hat{\vect{\eta}}^{(q)} \! = \! [\hat{\eta}_\lambda^{(q)}, \hat{\eta}_{\mathcal{I}, \scriptscriptstyle{0}}^{(q)}, \hat{\eta}_{\mathcal{I}, \tau}^{(q)}, \hat{\eta}_{p, \scriptscriptstyle{0}}^{(q)}, \hat{\eta}_{p, d}^{(q)},   \hat{\eta}_{p, v}^{(q)}]}$, where $\hat{\eta}_\lambda^{(q)}$ is the decoded length, and $\hat{\eta}_{\mathcal{I}}^{(q)}$ and $\hat{\eta}_{p}^{(q)}$ contain the respective coefficients for $f_{\mathcal{I}}$ and $f_p$, we model $q$ stochastically as
\begin{equation}
\begin{aligned}
    P(\mathcal{I}_q(t) = 1) &= f_{\mathcal{I}}(t; \hat{\eta}_{\mathcal{I}}^{(q)}), \\
    P(p_q(t) = s) &= f_p(s; t, \hat{\eta}_{p}^{(q)}). %\\
    %P(o_q(s, t) = 1) &= \hat{o}_q(s, t).
\end{aligned}\label{eq:vv_occ}
\end{equation}
Next, we derive the unimodal occupancy map ${\hat{o}_q(s, t)}$, referred to as the probabilistic occupancy \textit{footprint} of a single virtual vehicle $q$. We assume that $\mathcal{I}_q(t)$ and $p_q(t)$ are independent random variables and use the shorthand notation $s_{\pm} = s \pm \hat{\eta}_\lambda^{(q)}\! / \, 2$ to denote the occupancy bounds for a given $s$. As is illustrated in~\cref{fig:occpred}, applying \cref{eq:vv_occ_def} and~\cref{eq:vv_occ} results in
\begin{equation}
\hspace{-0.28cm}
\begin{aligned}
    \hat{o}_q(s, t) &= P(o_q(s, t) = 1) \\
    &= P(\mathcal{I}_q(t) = 1) P(|s - p_q(t)| < \hat{\eta}_\lambda^{(q)}\! / \, 2) \\
    &= f_{\mathcal{I}}(t; \hat{\eta}_{\mathcal{I}}^{(q)}) \! \int_{s_{-}}^{s_{+}} \!\!\! f_p(s'; t, \hat{\eta}_{p}^{(q)}) \d s', \\
    &= f_{\mathcal{I}}(t; \hat{\eta}_{\mathcal{I}}^{(q)}) 
    (F_p(s_{+}; t, \hat{\eta}_{p}^{(q)}){-}F_p(s_{-}; t, \hat{\eta}_{p}^{(q)})).
\end{aligned}\label{eq:occ_integral}
\end{equation} 
\begin{figure}[!ht]%
    {
    \vspace*{-0.1cm}\scalebox{0.9}[0.9]{\begin{overpic}[width=9cm,tics=10, trim=13 28 0 86,clip]{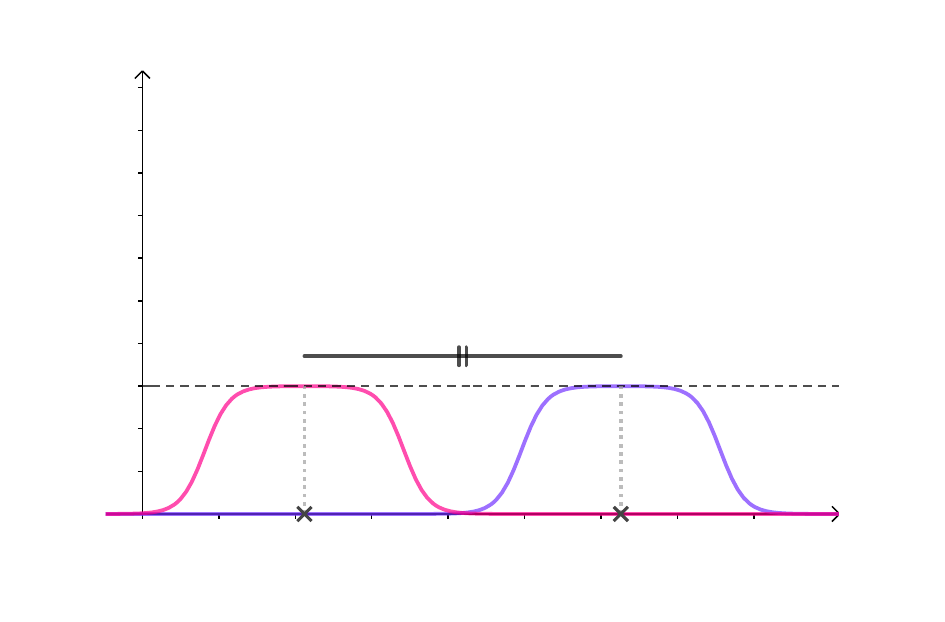}
 \put(9.33,25.64){
         \includegraphics[width=0.07cm, angle=0]{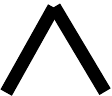}}
 \put (89,0.4) {\normalsize$t$}
 \put (5,19.3) {\scriptsize$0.4$}
 \put (5,9.85) {\scriptsize$0.2$}
 \put (37.7,21.0) {\small$\hat{\eta}_{\mathcal{I}, \tau}^{(2)} - \hat{\eta}_{\mathcal{I}, \tau}^{(1)}$}
 \put (81.7,17.1) {\small$\hat{\eta}_{\mathcal{I}, \scriptscriptstyle{0}}$}
 \put (12.1,9.1)  {\color{purple1} \normalsize$f_{\mathcal{I}}^{(1)}$}
 \put (47.2,9.1)  {\color{purple2} \normalsize$f_{\mathcal{I}}^{(2)}$}
\end{overpic}
    }}%
 \caption{Existence probability $f_{\mathcal{I}}(t)$ for two virtual vehicles with identical baseline existence probability $\hat{\eta}_{\mathcal{I}, \scriptscriptstyle{0}} = 0.3$ and different temporal offset parameters $(\hat{\eta}_{\mathcal{I}, \tau}^{(1)}, \hat{\eta}_{\mathcal{I}, \tau}^{(2)})$.}%
\label{fig:temporal_shifting}
\end{figure}
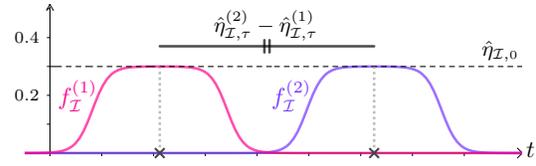 \vspace*{-0.5cm}
\subsection{Joint occupancy probability}
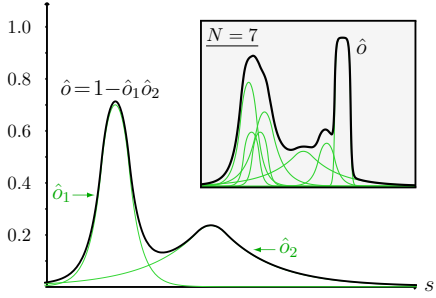
\begin{figure}[h!]%
    {    
    \vspace{0.1 cm}\scalebox{0.95}[0.95]{\resizebox{1.0\totalheight}{!}{
    \begin{overpic}[width=7cm,tics=10]{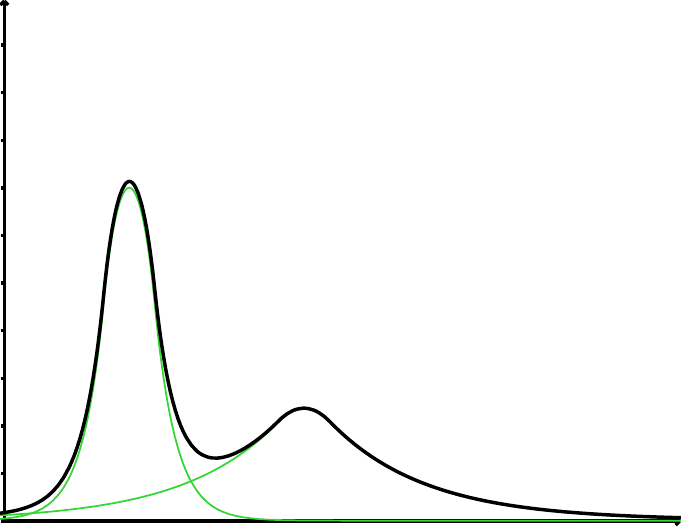}
 \put (102,0) {\large$s$}
 \put (4.3,52.6) {\large$\hat{o}\!=\!1{-}\hat{o}_1 \hat{o}_2$}
 \put (2,24.7) {\color{darkkkgreen} \large$\hat{o}_1$}
 \put(7.5,25.7){\color{darkkkgreen}\vector(1, 0){6}}
 \put (63.0,10.0) {\color{darkkkgreen} \large$\hat{o}_2$}
 \put(62.0,11.0){\color{darkkkgreen}\vector(-1, 0){6}}
 \put (-10,13.3) {\normalsize$0.2$}
 \put (-10,27.3) {\normalsize$0.4$}
 \put (-10,41.3) {\normalsize$0.6$}
 \put (-10,55.3) {\normalsize$0.8$}
 \put (-10,69.3) {\normalsize$1.0$}
 \put(40,28){
    \fbox{\includegraphics[width=4.0cm]{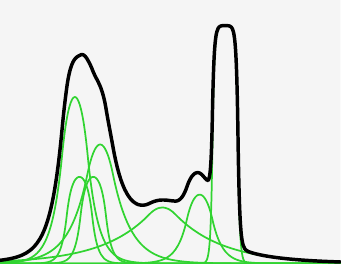}}}
 \put (83.7,63.6) {\large$\hat{o}$}
 \put (43.5, 67) {\uline{\normalsize$N = 7$}}
\end{overpic} }
    }%
    }
 \caption{Spatial cross-section of the probabilistic occupancy map $\hat{o}(s, t)$ induced by two virtual vehicles at a given time $t$. In the top right subplot, more virtual vehicles are added for comparison.}%
    \label{fig:occpred_joint}%
\end{figure}

We next derive an expression for the predicted joint occupancy $\hat{o}(s, t)$. To facilitate tractable inference, we assume the virtual vehicles to be independent. For a set of virtual vehicles ${\mathcal{Q} = \{q_1, ... , q_N\}}$, the probability of \textit{at least one} vehicle occupying $\Gamma_{\mathrm{ego}}$ at the location $[s, t]$ is given by
\begin{equation}
    \hat{o}(s, t) = 1 - \smashoperator{\prod_{q \in \mathcal{Q}}} \left(1 - \hat{o}_q(s, t)\right),\label{eq:jointoccpred}
\end{equation}
as exemplified in \cref{fig:occpred_joint}. Given a sufficiently large $N$, neither the independence assumption nor the simplified linear motion model assumed in~\cref{eq:fokker_sol_pdf} pose significant restrictions on the modeling capacity of our decoder, as~\cref{eq:jointoccpred} allows complex behavior patterns and the corresponding occupancy maps to be modeled via superimposed virtual vehicles.
\setlength{\fboxrule}{1pt}
\setlength{\fboxsep}{0pt}
\subsubsection{Recurrent decoding}
We employ a recurrent neural network $\Theta_{D}$ and a downscaling layer $\Theta_{q}$ to decode the virtual vehicle parameterizations $\{\hat{\vect{\eta}}^{(1)}, ... , \hat{\vect{\eta}}^{(N)}\}$ from $\vect{z}_{\mathrm{ego}}$. We transform the raw outputs of $\Theta_{q}$ via rescaled sigmoid functions so as to conform to ${\vect{\eta}_{\mathrm{min}} \leq \hat{\vect{\eta}}^{(q)} \hspace{-2pt} \leq \vect{\eta}_{\mathrm{max}}}$. The parameter bounds $\vect{\eta}_{\mathrm{min}}$ and $\vect{\eta}_{\mathrm{max}}$ correspond to reasonable priors for plausible vehicle behavior in traffic and are assumed to be fixed hyperparameters.
\subsection{Spatio-temporal occupancy loss}
\begin{figure}[htb!]%
    \centering
    \vspace{0.05cm}\hspace*{-0.35cm}\subfloat[$t= \SI{0.0}{\second}$.
    \label{fig:intbounds1}]{
    \resizebox{1.055\totalheight}{!}{
    \begin{overpic}[angle=-90, width=0.8\linewidth,tics=10]{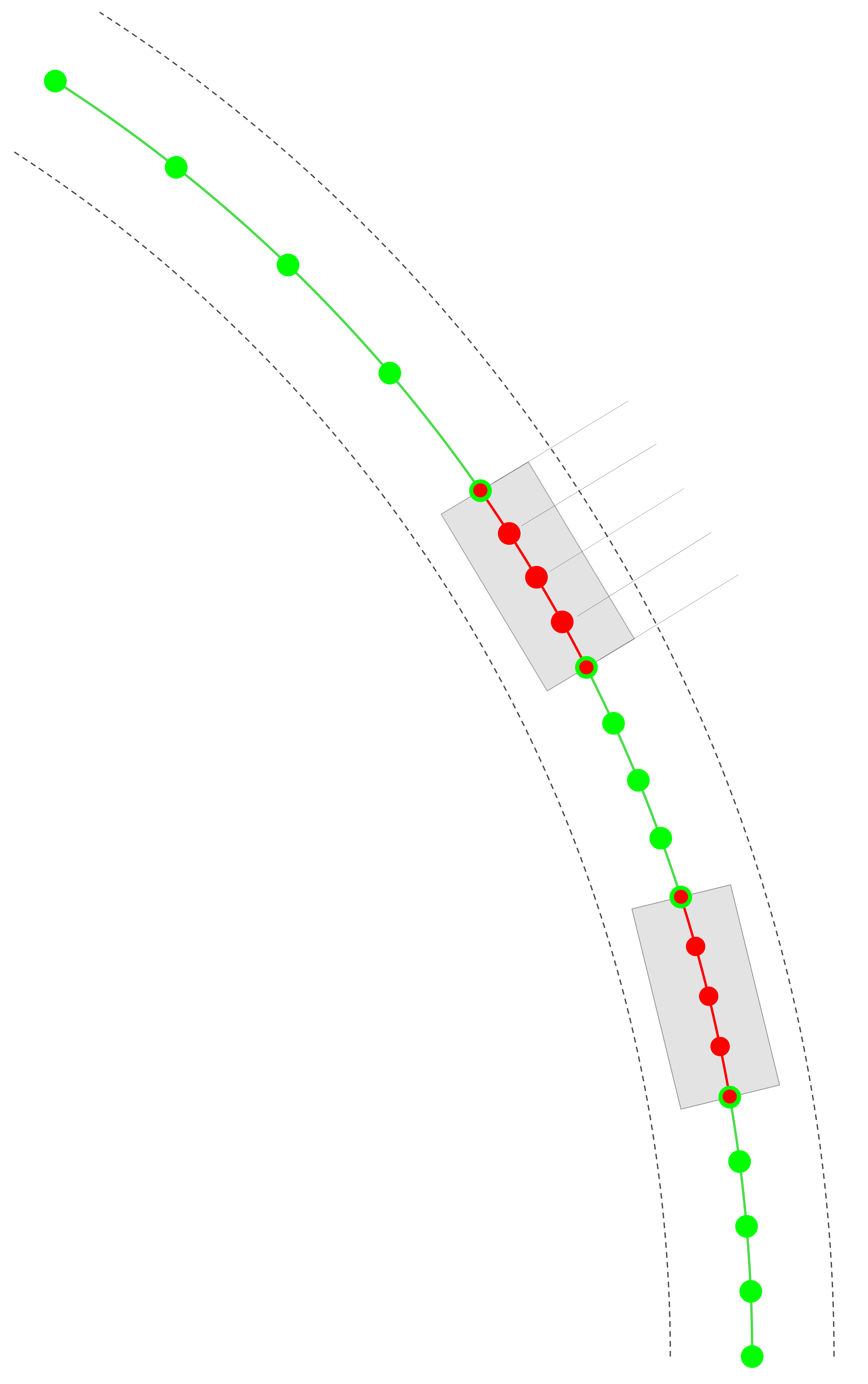}
 \put (21,32.5) {\large{$\mathcal{O}_n$}}
 \put (35,32.5) {\large{$\mathcal{O}_p$}}
 \put (3.8,57.9) {\color{black}\large{$\Omega_1$}}
 \put (12.9,57.9) {\color{black}\large{$\Omega_2$}}
 \put (22.0,57.9) {\color{black}\large{$\Omega_3$}}
 \put (31.1,57.9) {\color{black}\large{$\Omega_4$}}
 \put (49.2,57.9) {\color{black}\large{$\Omega_5$}}
 \put (67.8,54.0) {\color{black}\normalsize{$s$}}
 \put(16.2,53.5){\color{red}\vector(6, -5){19.0}}
 \put(35.4,53.5){\color{red}\vector(1, -5){3.05}}
 \put(6.5,53.5){\color{darkkgreen}\vector(4, -5){13.75}}
 \put(25.8,53.5){\color{darkkgreen}\vector(-1, -8){1.95}}
 \put(52.1,53.5){\color{darkkgreen}\vector(-16, -11){24.8}}
 \put (57.6,2.9) {\rotatebox{33}{\large{$R_{\ell} = 5$}}}
 \put(0,53){
         \includegraphics[width=4.5cm]{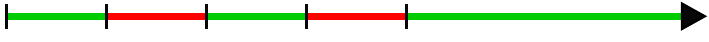}}
 \put(90.7, 56.6){
         \includegraphics[width=0.2cm, angle=45]{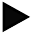}}
\end{overpic} 
}
    }%
    \hspace*{-0.5cm}\subfloat[$t= \SI{0.1}{\second}$.
    \label{fig:intbounds2}]{
    {
    \resizebox{1.055\totalheight}{!}{
    \begin{overpic}[angle=-90, width=0.8\linewidth,tics=10]{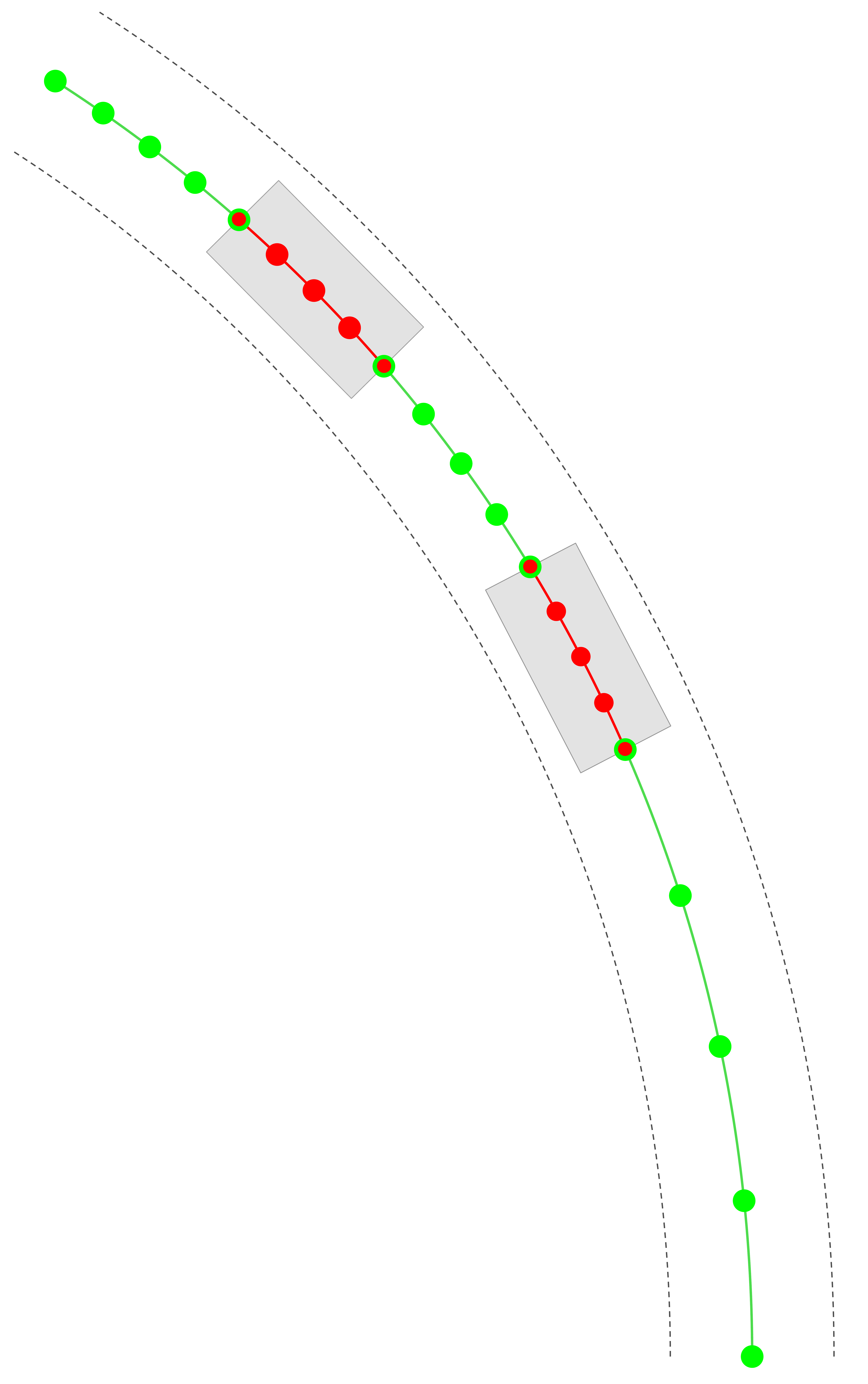}

 \put (21,32.5) {\large{$\mathcal{O}_n$}}
 \put (35,32.5) {\large{$\mathcal{O}_p$}}
 \put (12.9,57.9) {\color{black}\large{$\Omega_1$}}
 \put (31.65,57.9) {\color{black}\large{$\Omega_2$}}
 \put (40.425,57.9) {\color{black}\large{$\Omega_3$}}
 \put (49.8,57.9) {\color{black}\large{$\Omega_4$}}
 \put (58.2,57.9) {\color{black}\large{$\Omega_5$}}
 \put (67.8,54.0) {\color{black}\normalsize{$s$}}
 \put(34.4,53.5){\color{red}\vector(2, -13){2.35}}
 \put(53.4,53.5){\color{red}\vector(-4, -5){12.8}}
 \put(15.4,53.5){\color{darkkgreen}\vector(5, -11){7.0}}
 \put(43.5,53.5){\color{darkkgreen}\vector(-9, -8){17}}
 \put(61,53.5){\color{darkkgreen}\vector(-20, -11){31.4}}
 \put(0,53){
         \includegraphics[width=4.5cm]{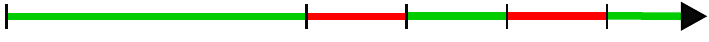}}
 \put(90.7, 56.6){
         \includegraphics[width=0.2cm, angle=45]{figures/arrow.pdf}}
\end{overpic}  
}
    }%
    }
 \caption{Loss evaluation for two subsequent time instances. The segment-wise integrals over the path regions in $\mathcal{O}_p^{(t)}$ and $\mathcal{O}_n^{(t)}$ are approximated numerically based on $R_{\ell}$ evenly spaced samples. %For illustrative purposes, we have set $R_{\ell} = 5$.
 }
    \label{fig:intbounds}%
\end{figure}

A naive approach towards defining the loss function is to consider the binary cross-entropy over a uniform grid discretization of the spatio-temporal occupancy domain~\cite{bender2022practical}. Instead, we compute the loss segment-wise in a boundary-aware fashion by leveraging the continuous output domain of our occupancy decoder. As shown in~\cref{fig:intbounds}, we let $\mathcal{O}_p^{(t)}$ and $\mathcal{O}_n^{(t)}$ contain the occupied and non-occupied connected segments of $\Gamma_{\mathrm{ego}}$ at time $t$. The respective path segments are indexed as $\Omega_{\square} \subseteq [0, \zeta_{\mathrm{ego}}]$ according to their topological order. Minimizing the binary cross-entropy, we then define our decoding loss as
\begin{align}
\nonumber \ell_p(t) = & - \!\!\!\!\!\! \sum_{\Omega \in \mathcal{O}_p^{(t)}} \frac{1}{|\Omega|} \int_{s \in \Omega} \log{(\hat{o}(s, t))} \, \d s, \\
\nonumber \ell_n(t) = & - \!\!\!\!\!\! \sum_{\Omega \in {\mathcal{O}_n^{(t)}}} \frac{1}{|\Omega|} \int_{s \in \Omega} \log{(1 - \hat{o}(s, t))} \, \d s, \\
\ell = & \int_{t=0}^{T} \delta_\ell^t \big( \ell_p(t) + \ell_n(t) \big) \, dt,
\end{align}\label{eq:lossfunction}
where $T \in \mathbb{R}$ is the considered time horizon and ${\delta_\ell \in [0, 1]}$ is a discount factor for reflecting the diminishing significance of future occupancy. With $|\Omega|$ denoting the arclength of the respective path segment, the normalization factor ${1 / |\Omega|}$ is introduced to avoid the dependence on segment length and to counteract the class imbalance caused by road surfaces being predominantly unoccupied.
As the spatial integrals are otherwise intractable, we approximate them numerically with resolution $R_{\ell}$ as illustrated in~\cref{fig:intbounds1}. Similarly, we discretize the temporal integral over $T_D$ time steps.

\section{Numerical experiments}\label{sec:implementation}

Next, we describe the collection of our simulated traffic dataset and introduce the numerical RL experiments that evaluate the effectiveness of our learned state representations.

\subsection{Dataset}
Using the OpenStreetMap~\cite{OpenStreetMap} API, our dataset is comprised of a diverse set of urban locations sampled from within the Munich metropolitan region. The collected road networks are then populated with vehicles using the traffic simulator SUMO~\cite{lopez_microscopic_2018}. In total, the dataset contains 1000 simulated scenarios, jointly amounting to 28 hours of traffic.
\subsection{Reinforcement learning agent}\label{sec:reinforcement}
For our experiments, we extract $\vect{z}_{\mathrm{ego}}$ as inputs for a PPO-based~\cite{PPO} RL agent together with the current ego velocity $v_{\mathrm{ego}}$. At each time step, the encoder is conditioned on the navigation context generated by a high-level route planner. The resulting reference routes span across multiple lanelets and include heterogeneous map structures such as intersections. We intentionally limit the agent to longitudinal acceleration control along the reference path to focus on the effect of the learned representations. 
Given the weighting coefficients ${\vect{w} \in \mathbb{R}^4}$, the reward ${r}$ is defined as ${{r} = \vect{w} \cdot [r_{path}, r_{collision}, r_{speed}, r_{\hat{o}}]}$, where $r_{path}$ is a dense path progression reward, $r_{collision}$ is a sparse penalty imposed on collisions, $r_{speed}$ is a linear over-speeding penalty, and $r_{\hat{o}}$ penalizes expected occupancy conflict from
\begin{equation}
    r_{\hat{o}} = \int_{t=0}^T \delta_{r}^t \int_{\tilde{s}(t)-\nicefrac{\lambda_{\mathrm{ego}}}{2}}^{\tilde{s}(t)+\nicefrac{\lambda_{\mathrm{ego}}}{2}} \hat{o}(s, t) \d s \d t,
\end{equation}\label{eq:reward_eq}
\noindent with ${\tilde{s}(t)=v_{\mathrm{ego}} t}$ being a constant-velocity extrapolation of the ego position and $\delta_{r} \in [0, 1]$ being a discount factor. 
\subsection{Baseline approaches}\label{sec:baselines}
We compare our PPO agent's performance against multiple baselines trained with identical reward configurations:
\begin{enumerate}
    \item \textbf{V2V}: A vehicle-to-vehicle (i.e., not map-aware) GNN policy network as proposed in~\cite{hart_graph_2020}.
    \item \textbf{V2L}: A direct adoption of our \textsc{Encoder} as policy network (i.e., without pre-training $\vect{z}_{\mathrm{ego}}$).
    \item \textbf{Naive}: The pre-trained representations resulting from using a feed-forward network $\mathrm{MLP}(256, 128)$ for decoding $\hat{o}(s, t)$ from ${[\vect{z}_{\mathrm{ego}}, s, t]}$ without architectural constraints. This naively assumes occupancy to be an independent property for each spatio-temporal coordinate vector $[s, t]$.
\end{enumerate}

\subsection{Implementation and training}
  The end-to-end training of the representation model was conducted on an NVIDIA A100 Tensor Core GPU for 48 hours using the Adam optimizer~\cite{ADAMOPT} with the hyperparameters listed in~\cref{tab:repr_hyperparams}. Subsets of the collected dataset were used for training (90\%) and testing (10\%). To aid generalization, random planning contexts for the encoder were resampled for each mini-batch during training. The PPO agents were implemented using the {Stable-Baselines 3} framework~\cite{stable-baselines3}  and trained for $10^6$ steps on simulated replays of the scenarios, with start and goal positions for the ego vehicle being randomly sampled for each episode.
 
 \begin{table}[!ht]
    \centering
    {\footnotesize
	\begin{tabular}{lll}
		\toprule
		Encoding layers ($\Theta_{\textsc{l}}, \Theta_{\textsc{v2l}}$, $\Theta_{\textsc{l2l}}$, $\Theta_{C}$, $\Theta_{z})$ & $\mathrm{Linear}$
		\\
		Encoding dimensions ($H$, $Z$) & 256, 32 \\
		Number of \textsc{l2l} layers (${L}$) & 4 \\
		Aggregation function ($\Sigma$) & $\mathrm{max}$
		\\
		Activation function ($\rho$) & $\mathrm{tanh}$
		\\
		Reference path length ($\zeta_{\mathrm{ego}}$) & \SI{45}{\meter} \\
		\midrule
		Decoding layers ($\Theta_{D}$, $\Theta_{q}$) & $\mathrm{LSTM}$(256), $\mathrm{Linear}$  
		\\
		Number of virtual vehicles ($N$) & 12   
		\\
		Temporal masking constants ($\tau_R, \tau_C$) & 6.0, 0.7 \\
		Decoding horizon ($T, T_D$) & \SI{2.4}{\second}, 60 \\
		Spatial integration method & $\mathrm{Trapezoidal}(R_{\ell} = 40)$ \\
		Temporal occupancy discount factor ($\delta_\ell$) & 0.99 \\
		\bottomrule
	\end{tabular}}
	\caption[Architectural hyperparameters.]{Selected hyperparameters for our experiments. We refer to our online available implementation for further details.
 }
	\label{tab:repr_hyperparams}
\end{table}\vspace{-0.4cm} 

\section{Results}\label{sec:results}
\begin{figure}
\centering
\vspace{0.2cm}\begin{subfigure}[!htb]{\linewidth}
  \centering   
  \hspace{0.5cm}\begin{overpic}[width=\linewidth, trim=0 0 0 160,clip]{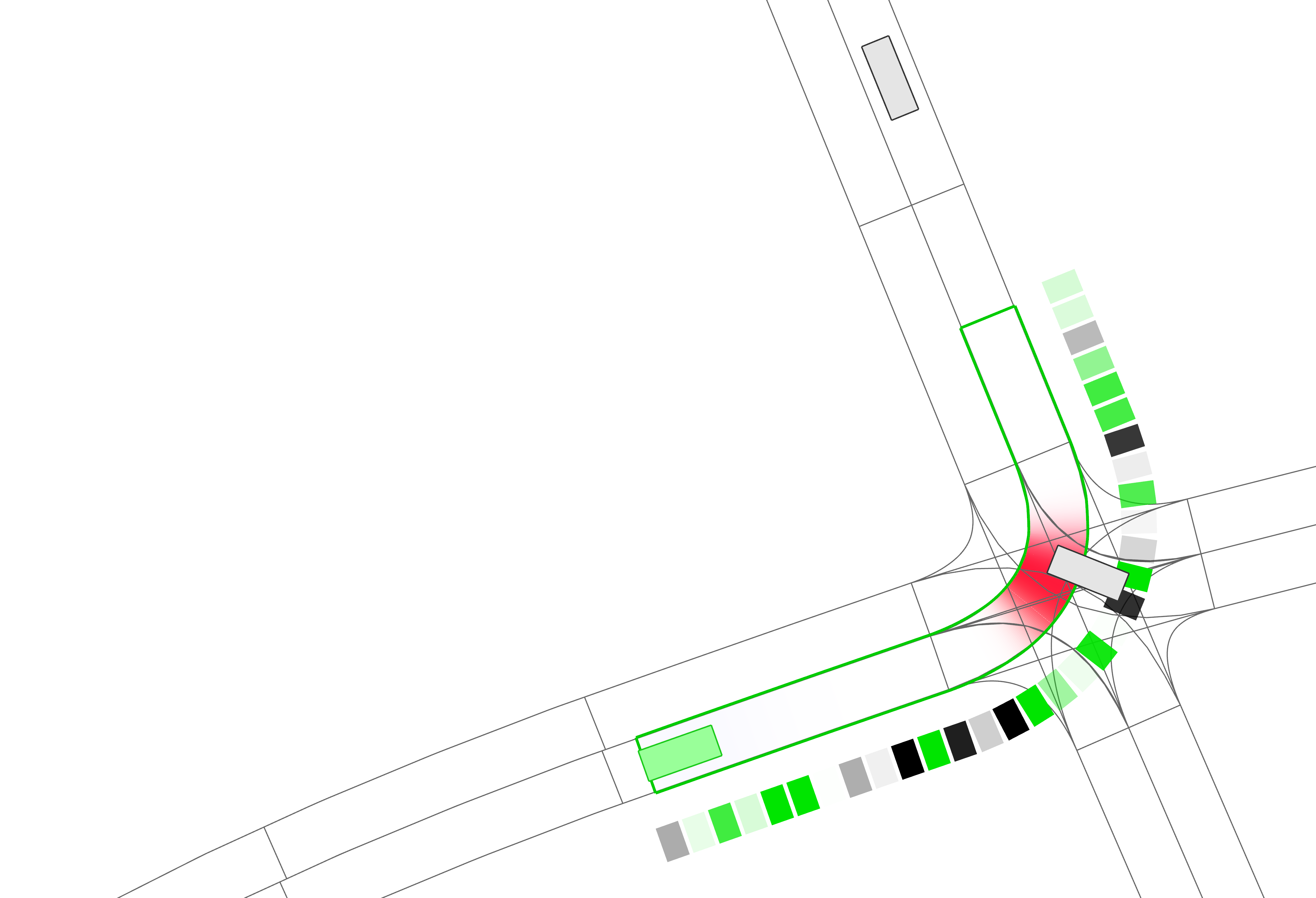}
         \put(-3,16){
         \includegraphics[width=4.2cm]{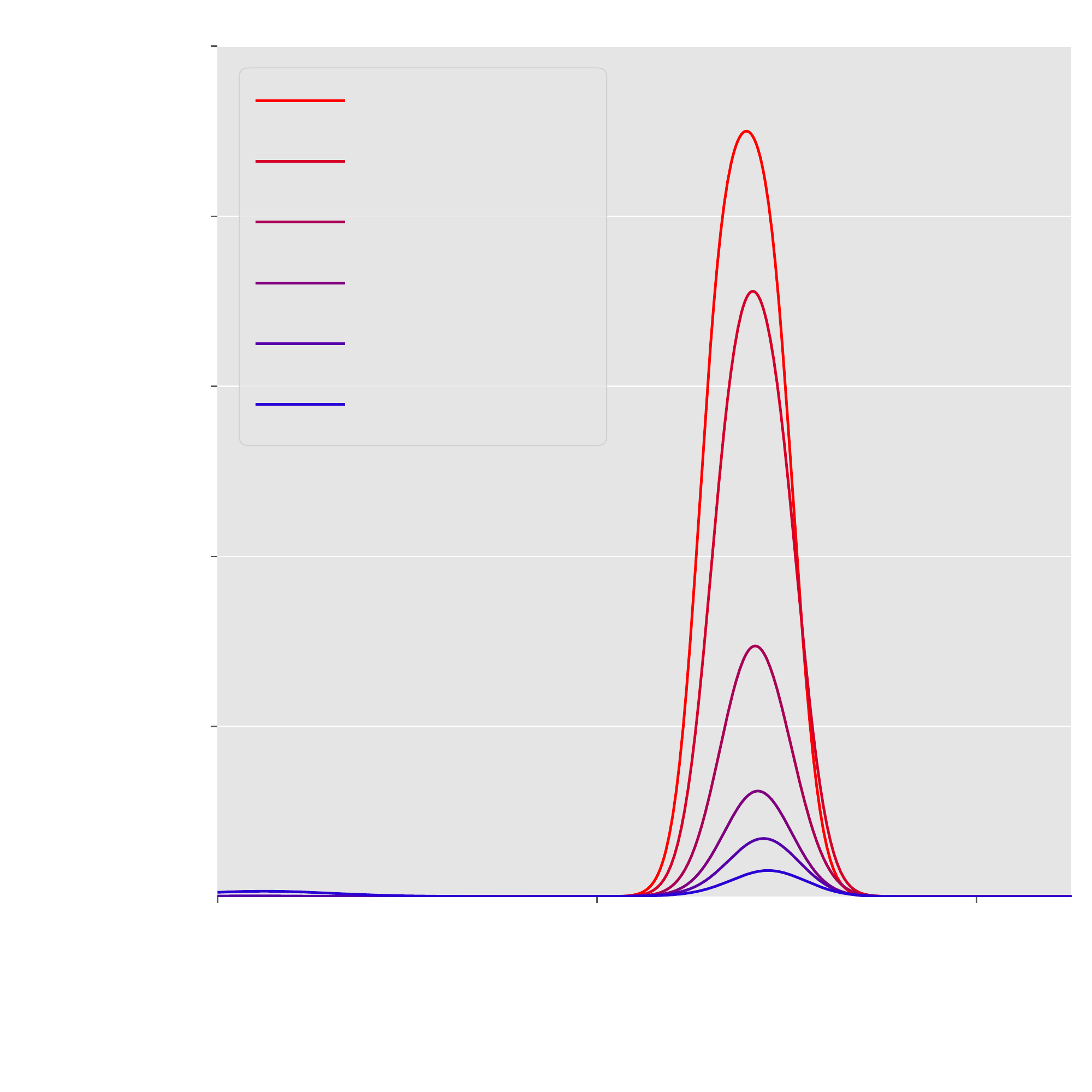}}
         \put(67,-4){
         \includegraphics[width=0.88cm]{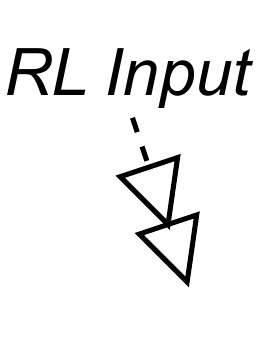}}
 \put (3.2,31.5) {\color{gray} {\scriptsize $0.2$}}
 \put (3.2,39.0) {\color{gray} {\scriptsize $0.4$}}
 \put (3.2,46.5) {\color{gray} {\scriptsize $0.6$}}
 \put (3.2,54) {\color{gray} {\scriptsize $0.8$}}
 \put (3.2,61.5) {\color{gray} {\scriptsize $1.0$}}
 \put (23,18.5) {\color{gray} {\normalsize $s \; [m]$}}
 \put (7,22.0) {\color{gray} {\scriptsize $0$}}
 \put (23,22.0) {\color{gray} {\scriptsize $20$}}
 \put (40, 22.0) {\color{gray} {\scriptsize $40$}}
 \put (-1.5,39.5) {\color{gray} {\normalsize \rotatebox{90} {$\hat{o}(s, t)$} }}
 \put (15,59.5) {\color{gray} {\tiny $t \! = \! 0.0 \; \SI{}{\second}$}}
 \put (15,56.8) {\color{gray} {\tiny $t \! = \! 0.4 \; \SI{}{\second}$}}
 \put (15,54.1) {\color{gray} {\tiny $t \! = \! 0.8 \; \SI{}{\second}$}}
 \put (15,51.4) {\color{gray} {\tiny $t \! = \! 1.2 \; \SI{}{\second}$}}
 \put (15,48.7) {\color{gray} {\tiny $t \! = \! 1.6 \; \SI{}{\second}$}}
 \put (15,46) {\color{gray} {\tiny $t \! = \! 2.0 \; \SI{}{\second}$}}
  \end{overpic}
  \caption{Vehicle from another lane intersecting $\Gamma_{\mathrm{ego}}$. The occupancy probability map has a temporary peak located at the intersection point.}
\end{subfigure}\vspace{0.2cm}
\begin{subfigure}[!htb]{\linewidth}
  \centering   
  \hspace{0.5cm}\begin{overpic}[width=\linewidth, trim=0 60 10 200,clip]{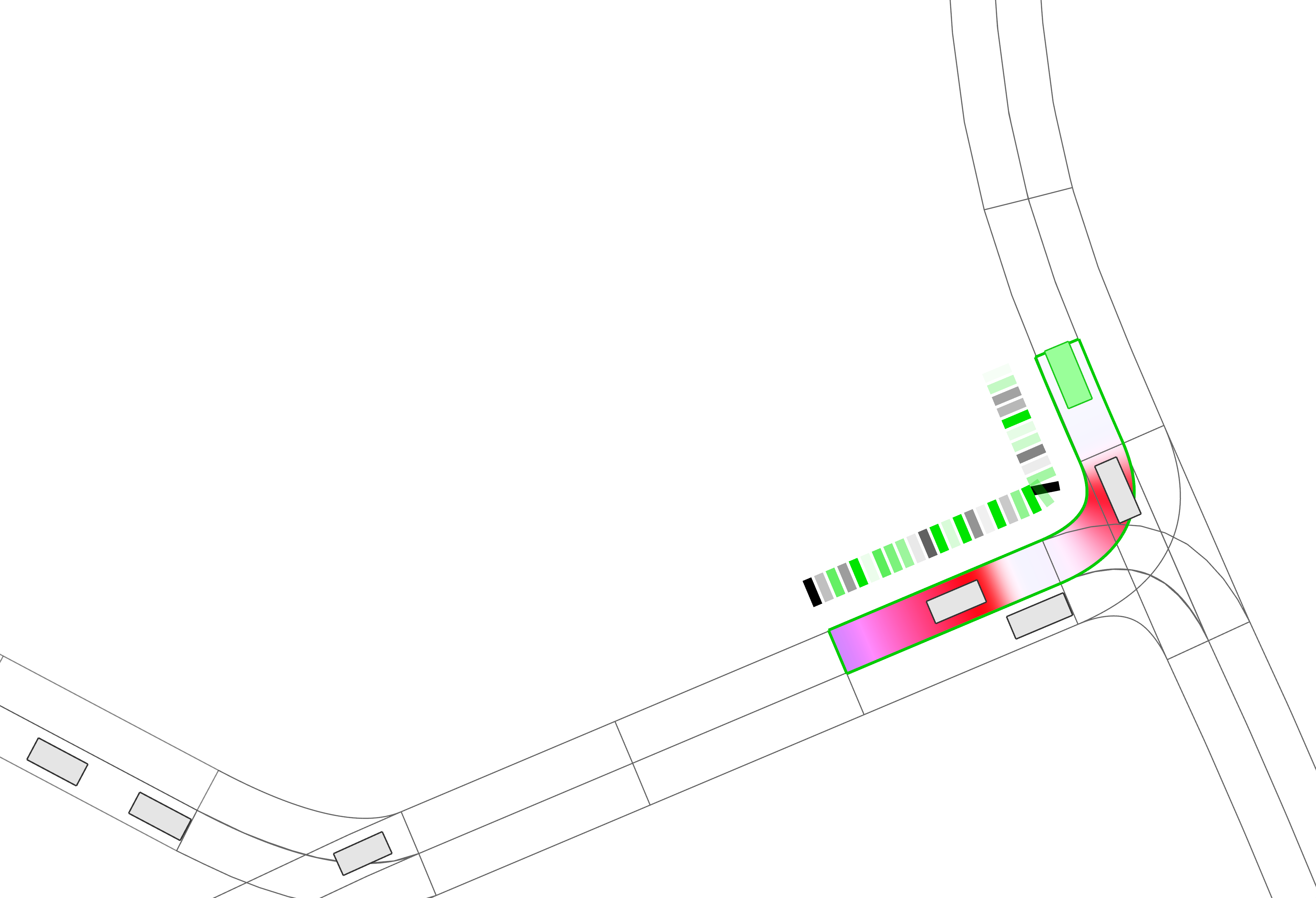}
         \put(-3,20.6){
         \includegraphics[width=4.2cm]{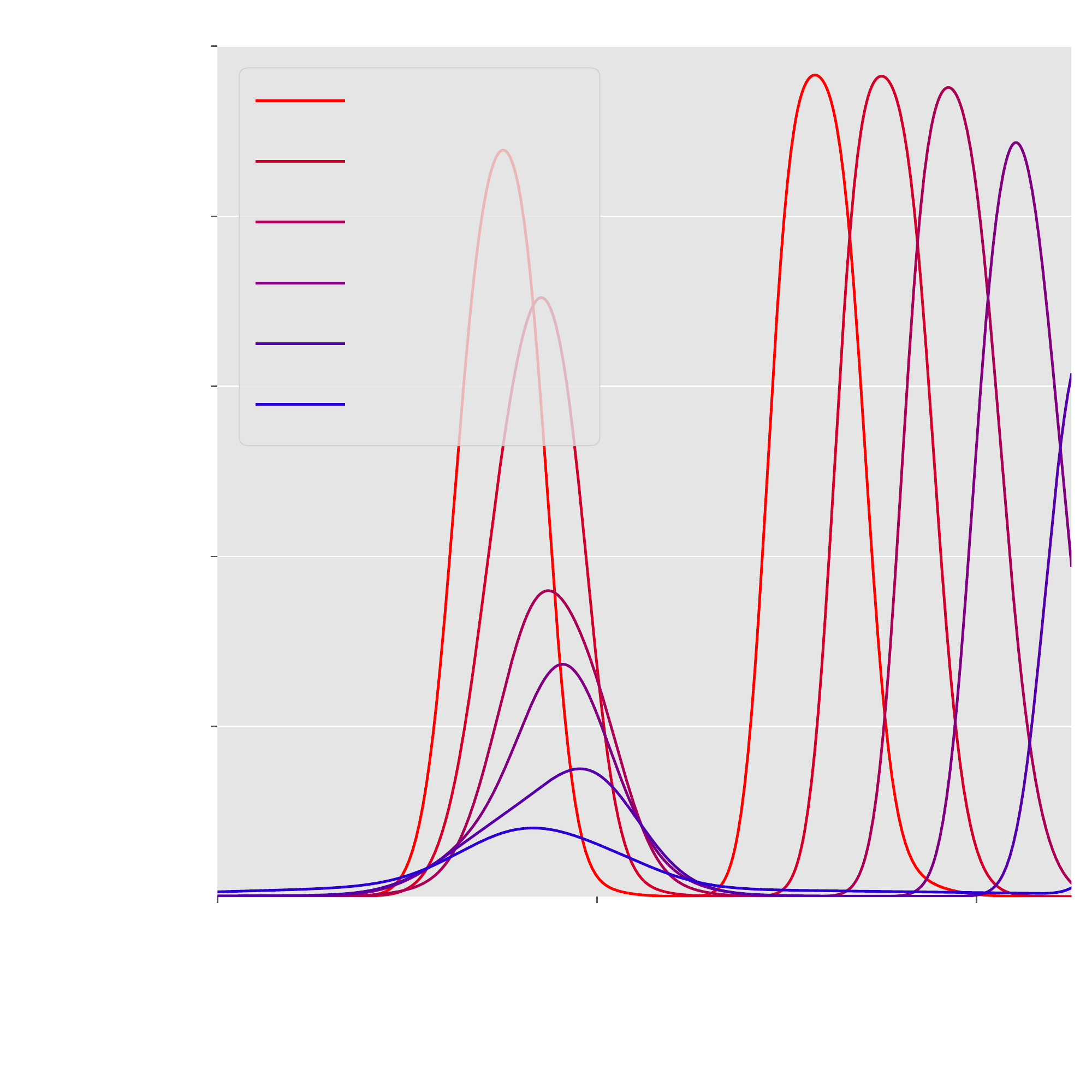}}
         \put(58,33.5){
         \includegraphics[width=0.9cm]{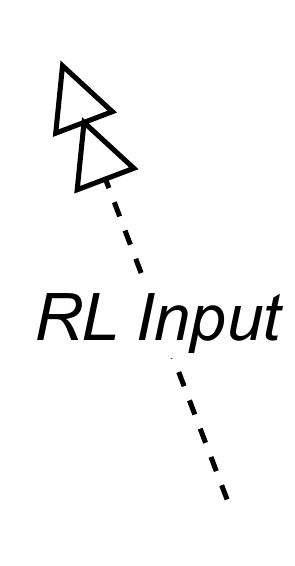}}
         \put (71.4,9.6) {\scriptsize$\displaystyle\color{black}{A}$}
         \put(73.0,12.6){\color{black}\vector(0, 1){6.8}}
         \put (92.8, 29.2) {\scriptsize$\displaystyle\color{black}{B}$}
         \put(92.8, 29.9){\color{black}\vector(-1, 0){6}}
         \put (3.2,31.5) {\color{gray} {\scriptsize $0.2$}}
 \put (3.2,39.0) {\color{gray} {\scriptsize $0.4$}}
 \put (3.2,46.5) {\color{gray} {\scriptsize $0.6$}}
 \put (3.2,54) {\color{gray} {\scriptsize $0.8$}}
 \put (3.2,61.5) {\color{gray} {\scriptsize $1.0$}}
 \put (23,18.0) {\color{gray} {\normalsize $s \; [m]$}}
 \put (7.2,21.7) {\color{gray} {\scriptsize $0$}}
 \put (23.5,21.7) {\color{gray} {\scriptsize $20$}}
 \put (41, 21.7) {\color{gray} {\scriptsize $40$}}
 \put (-1.5,39.5) {\color{gray} {\normalsize \rotatebox{90} {$\hat{o}(s, t)$} }}
 \put (15,59.5) {\color{gray} {\tiny $t \! = \! 0.0 \; \SI{}{\second}$}}
 \put (15,56.8) {\color{gray} {\tiny $t \! = \! 0.4 \; \SI{}{\second}$}}
 \put (15,54.1) {\color{gray} {\tiny $t \! = \! 0.8 \; \SI{}{\second}$}}
 \put (15,51.4) {\color{gray} {\tiny $t \! = \! 1.2 \; \SI{}{\second}$}}
 \put (15,48.7) {\color{gray} {\tiny $t \! = \! 1.6 \; \SI{}{\second}$}}
 \put (15,46) {\color{gray} {\tiny $t \! = \! 2.0 \; \SI{}{\second}$}}         
  \end{overpic}
  \caption{The rapid intensity decay of the leftmost probability mode suggests an accurate representation of B's likely departure from the ego route.}
\end{subfigure}
\caption{Decoded occupancy maps $\hat{o}(s, t)$ visualized together with color encodings of the corresponding latent representations ${\vect{z}_{\mathrm{ego}}}$, which are extracted as state observations for the RL agent.
}
\label{fig:examples_occ}
\end{figure}
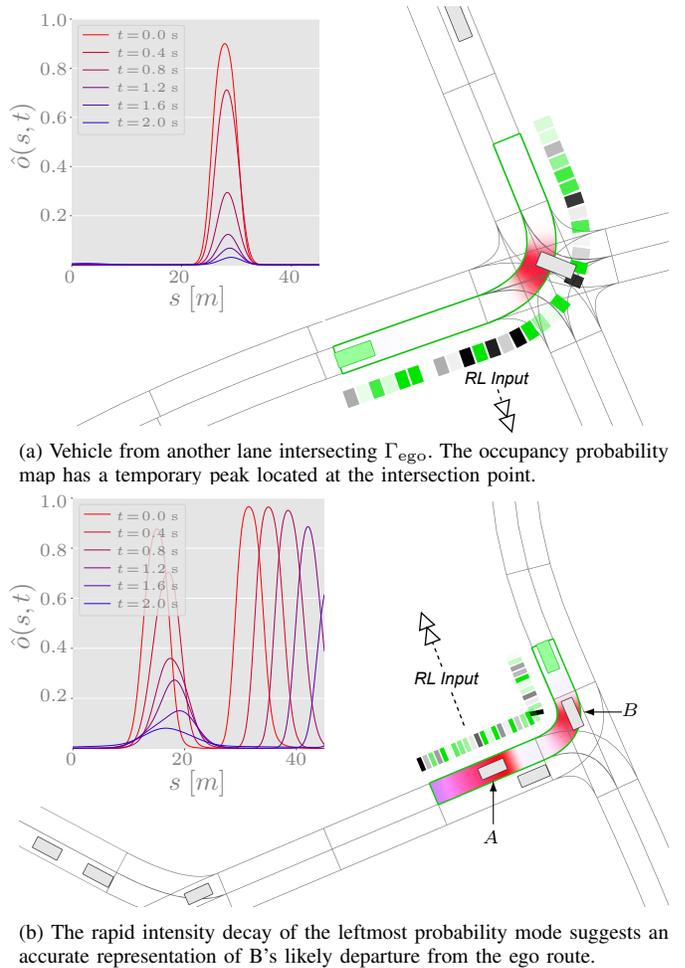

As is evident from the results reported in~\cref{tab:losstable}, our proposed model achieves better decoding performance than the unconstrained baseline. This indicates that our approach mitigates the effects of the information bottleneck caused by the encoding pipeline. Specifically, it is likely that the simpler hypothesis space streamlines the training process.
Further, a qualitative assessment of the decoded probability maps shown in~\cref{fig:examples_occ} suggests that our model is able to accurately predict complex environments to a degree where the intermediate encodings $\vect{z}_{\mathrm{ego}}$ will enable intelligent planning decisions for the agent.

\begin{table}[!htb]
    \footnotesize
    \caption{Empirical evaluation results on the test dataset.}\vspace{-0.1cm} 
    \begin{subtable}[t]{.5\linewidth}
      \centering
        \caption{Occupancy decoding loss.}
        \begin{tabular}{cc}
		\toprule
  model & $\ell$ \\ 
  \hline
  \addlinespace[1ex]
   Ours & \bf{1.045} \\
   Naive & 1.210 \\ 
  \bottomrule
\end{tabular}
    \end{subtable}%
    \begin{subtable}[t]{.5\linewidth}
      \centering
        \caption{Downstream RL performance.}
        \begin{tabular}{cc}
  \toprule
  agent & goal reach (\%) \\ 
  \hline
  \addlinespace[1ex]
   %OURS
   Ours & \bf{72.9} \\ 
   
   % HART
   V2V & 39.9 \\ 
   
   % HETRO
   V2L & 49.0 \\
   
   % MLP DECODER
   Naive & 54.0 \\
  \bottomrule
\end{tabular}
    \end{subtable}
\label{tab:losstable}
\end{table}\vspace{-0.02cm} 
The effectiveness of our representations is confirmed by the results of the conducted RL experiments. The improved success rate of the representation-enhanced agent indicates that the pre-trained representations simplify its motion planning task. Effectively, using them as state observations frees the agent from the responsibility of modeling its own surroundings, allowing it to concentrate its learning capacity on the lower-level control aspects of motion planning. As traffic modeling is a complex endeavour that is more easily tackled in a supervised setting, it is thus unsurprising that the simplification of the RL task improves the final performance.
\vspace{-0.2cm}
\section{Conclusion}\label{sec:conclusion}
We present a novel encoder-decoder architecture for learning latent state representations that enhance the performance of RL-based motion planning in heterogeneous driving environments. Our approach recurrently decodes and aggregates virtual vehicles to predict the spatio-temporal occupancy map. This ensures that the representation space is constrained to physically plausible predictions, which further enhances the model's interpretability and reduces its black-box nature. By using a heterogeneous GNN encoder to compress the traffic surroundings, our approach offers a significant benefit compared to previous works; as opposed to feature vectors narrowly tailored to specific traffic settings, our approach naturally extends to arbitrary road networks.

\section*{Acknowledgements}
This research was funded by the German Research Foundation grant AL 1185/7-1 and the German Federal Ministry for Digital and Transport through the project KoSi.

\interfootnotelinepenalty=10000
\AtNextBibliography{\footnotesize}
\balance
\printbibliography

\end{document}